\documentclass[runningheads]{llncs}

\usepackage{eccv}

\usepackage{eccvabbrv}

\usepackage{graphicx}
\usepackage{booktabs}
\usepackage{arydshln}
\usepackage[accsupp]{axessibility}  % Improves PDF readability for those with disabilities.

\usepackage[pagebackref,breaklinks,colorlinks,citecolor=eccvblue]{hyperref}
\usepackage{orcidlink}

\begin{document}

\title{Gradient-based Class Weighting for Unsupervised Domain Adaptation in Dense Prediction Visual Tasks}

\titlerunning{GBW}


\author{Roberto Alcover-Couso\inst{1} \and
Marcos Escudero-Viñolo\inst{1} \and
 Juan C. SanMiguel\inst{1} \and Jesus Bescos\inst{1}}

\institute{Universidad Autonoma de Madrid}

\maketitle

\begin{abstract}
  
In unsupervised domain adaptation (UDA), where models are trained on source data (e.g., synthetic) and adapted to target data (e.g., real-world) without target annotations, addressing the challenge of significant class imbalance remains an open issue. Despite considerable progress in bridging the domain gap, existing methods often experience performance degradation when confronted with highly imbalanced dense prediction visual tasks like semantic and panoptic segmentation. This discrepancy becomes especially pronounced due to the lack of equivalent priors between the source and target domains, turning class imbalanced techniques used for other areas (e.g., image classification) ineffective in UDA scenarios. This paper proposes a class-imbalance mitigation strategy that incorporates class-weights into the UDA learning losses, but with the novelty of estimating these weights dynamically through the loss gradient, defining a Gradient-based class weighting (GBW) learning. GBW naturally increases the contribution of classes whose learning is hindered by large-represented classes, and has the advantage of being able to automatically and quickly adapt to the iteration training outcomes, avoiding explicitly curricular learning patterns common in loss-weighing strategies. Extensive experimentation validates the effectiveness of GBW across architectures (convolutional and transformer), UDA strategies (adversarial, self-training and entropy minimization), tasks (semantic and panoptic segmentation), and datasets (GTA and Synthia). Analysing the source of advantage, GBW consistently increases the recall of low represented classes.%, as can be evaluated with the code provided at \href{https://github.com/}{hidden for anonymity}.
\end{abstract}

\begin{figure}[htp]
    \centering
    \begin{subfigure}[b]{0.246\linewidth}
    \caption{Color Image}
        \includegraphics[trim=1360 550 270 250,clip, width=\linewidth,]{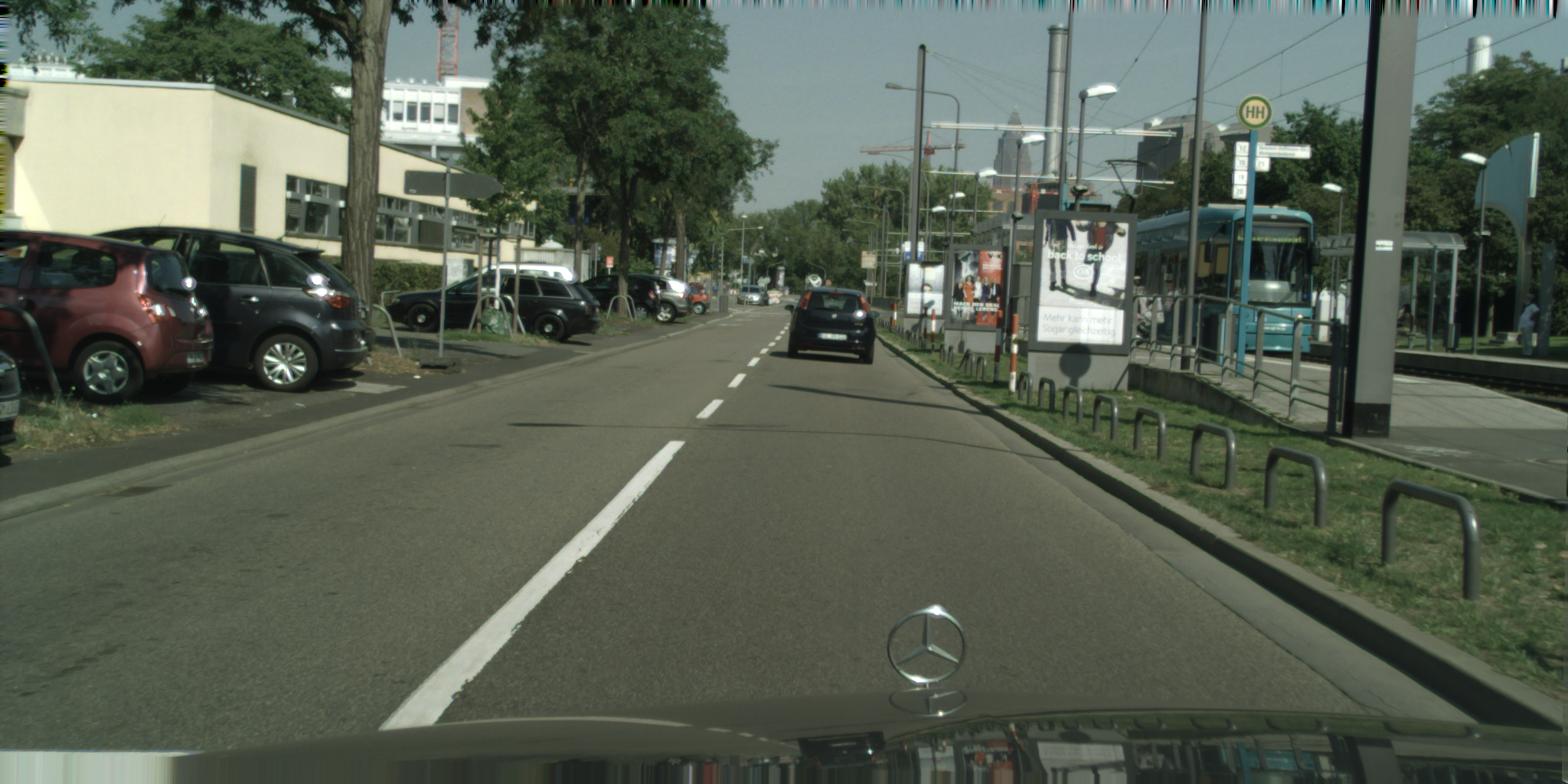}
    \end{subfigure}\hfill
    \begin{subfigure}[b]{0.246\linewidth}
    \caption{Ground truth}
        \includegraphics[trim=1360 550 270 250,clip, width=\linewidth,]{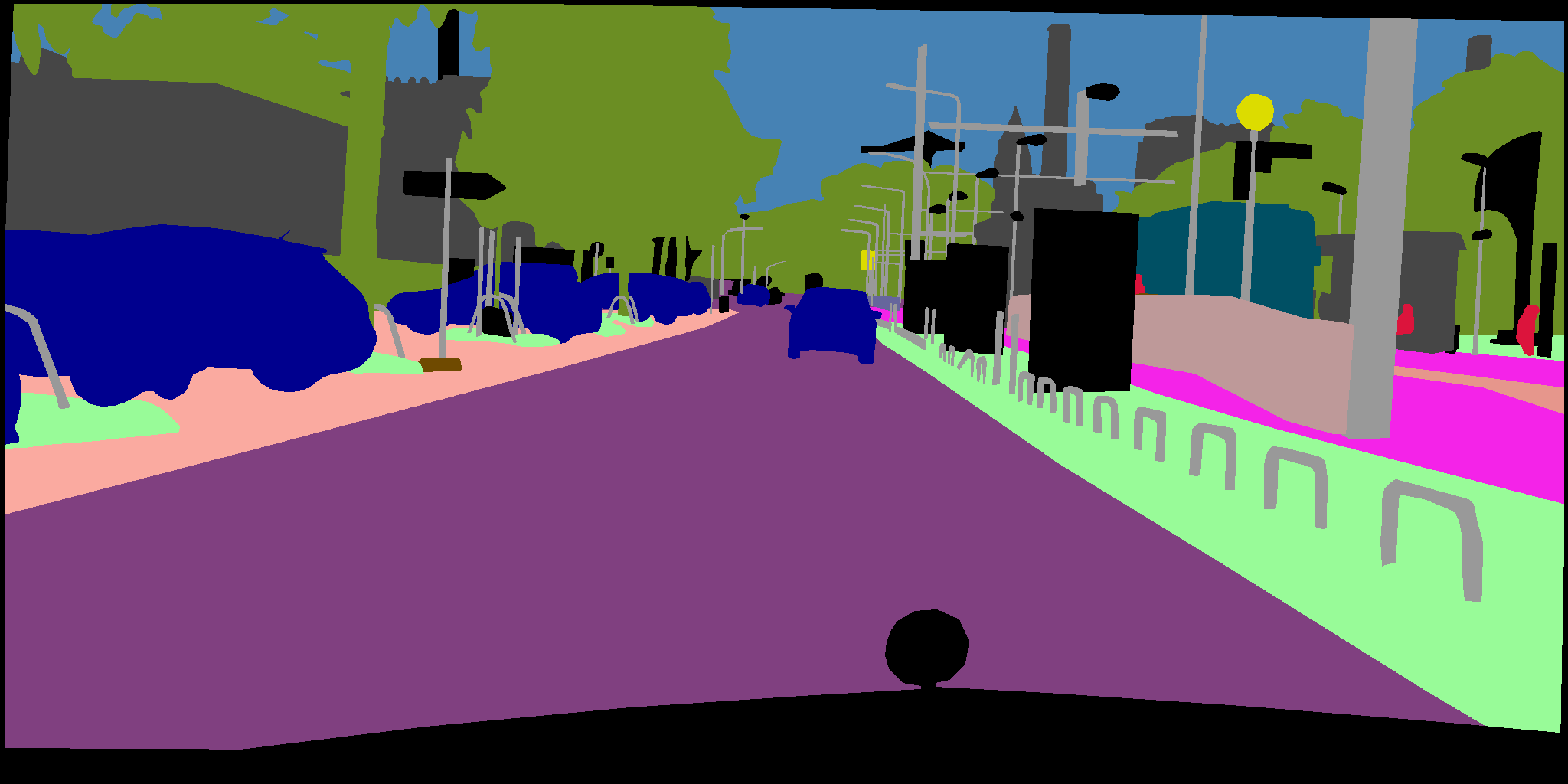}
    \end{subfigure}\hfill
    \begin{subfigure}[b]{0.246\linewidth}
    \caption{SOTA }
        \includegraphics[trim=1360 550 270 250,clip, width=\linewidth,]{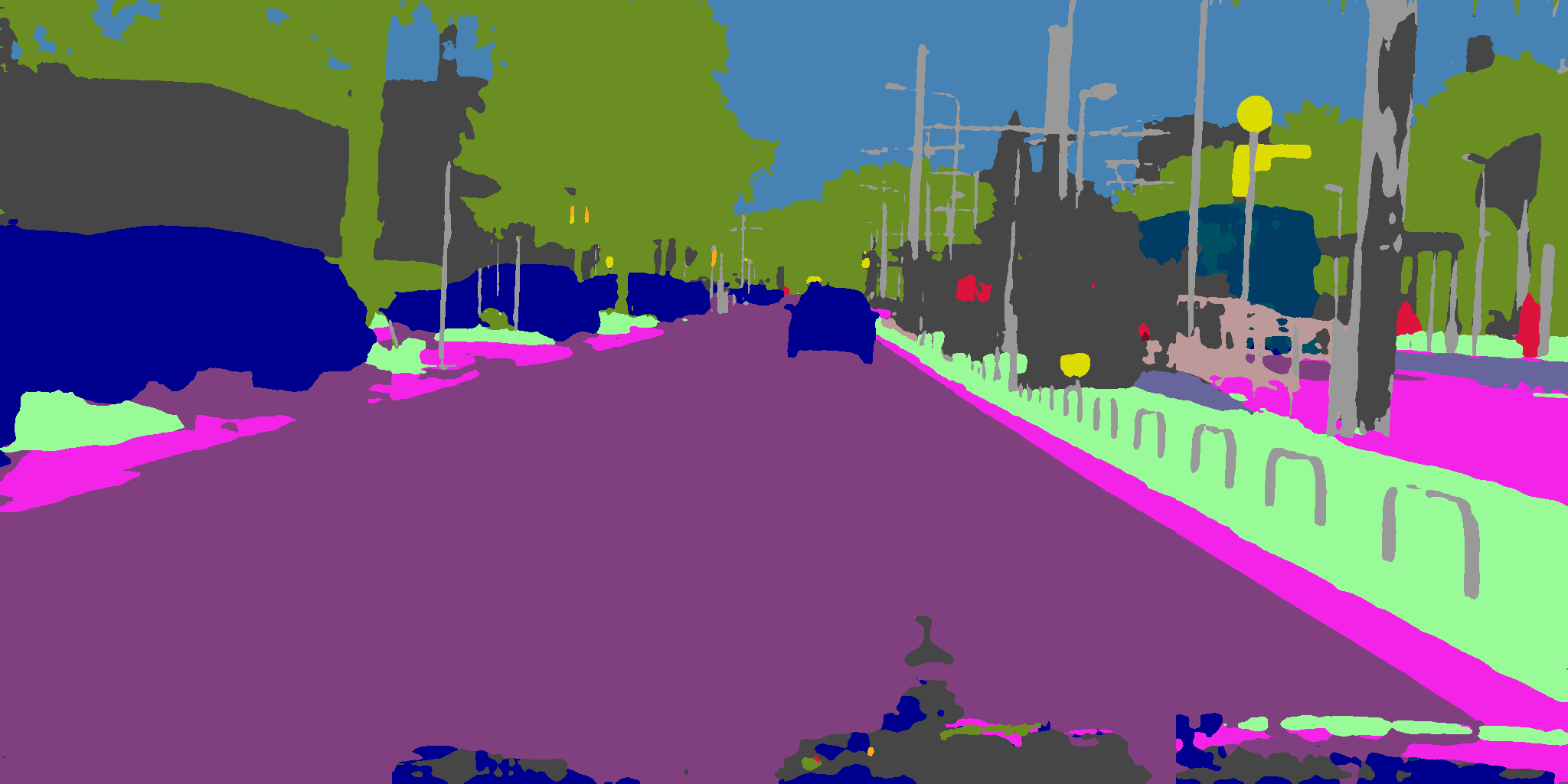}
    \end{subfigure}\hfill
    \begin{subfigure}[b]{0.246\linewidth}
    \caption{GBW}
        \includegraphics[trim=1360 550 270 250,clip, width=\linewidth,]
        {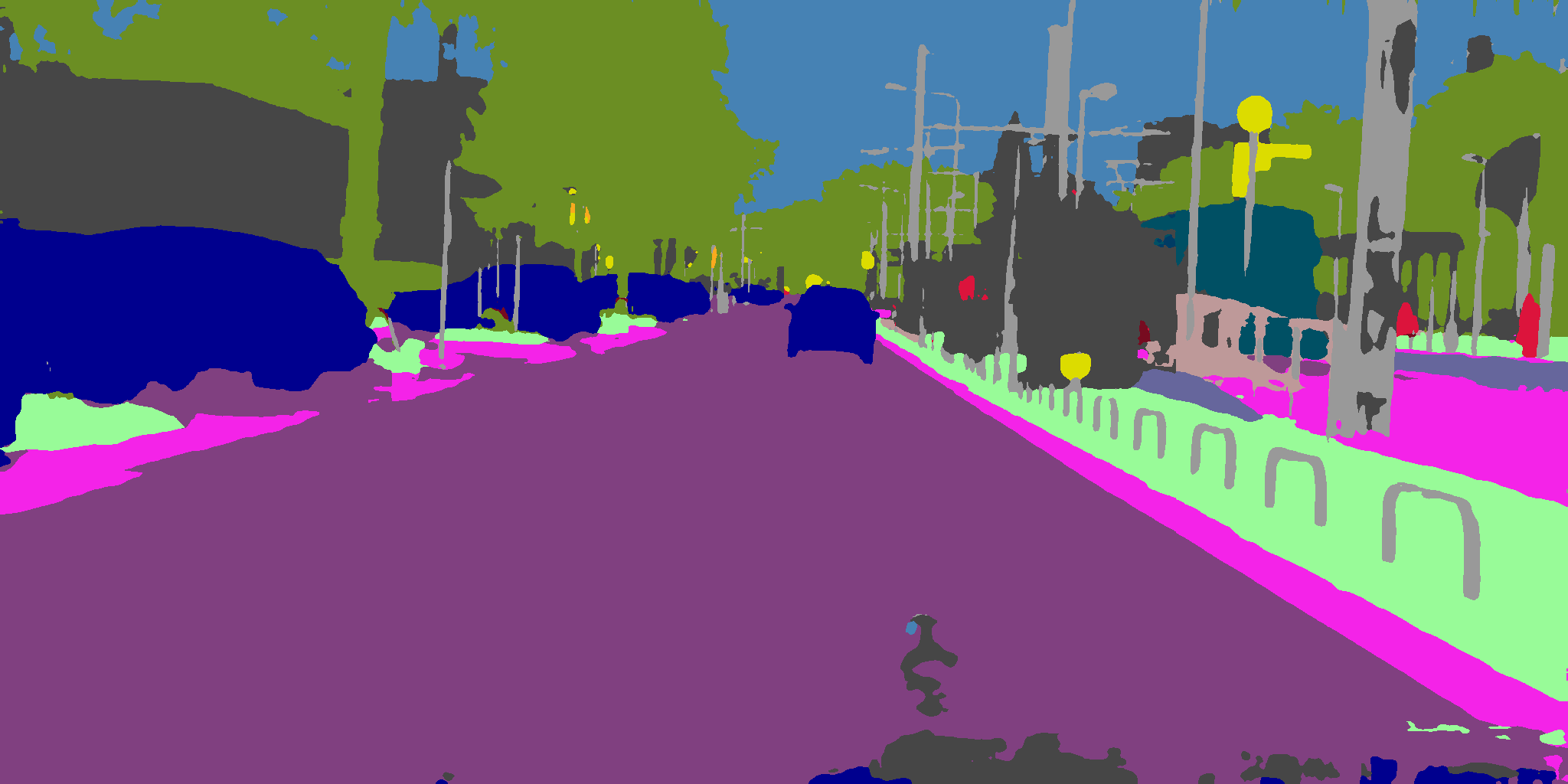}
    \end{subfigure}\\
    \begin{subfigure}[b]{0.246\linewidth}
        \includegraphics[trim=10 1300 870 100,clip, width=\linewidth,]{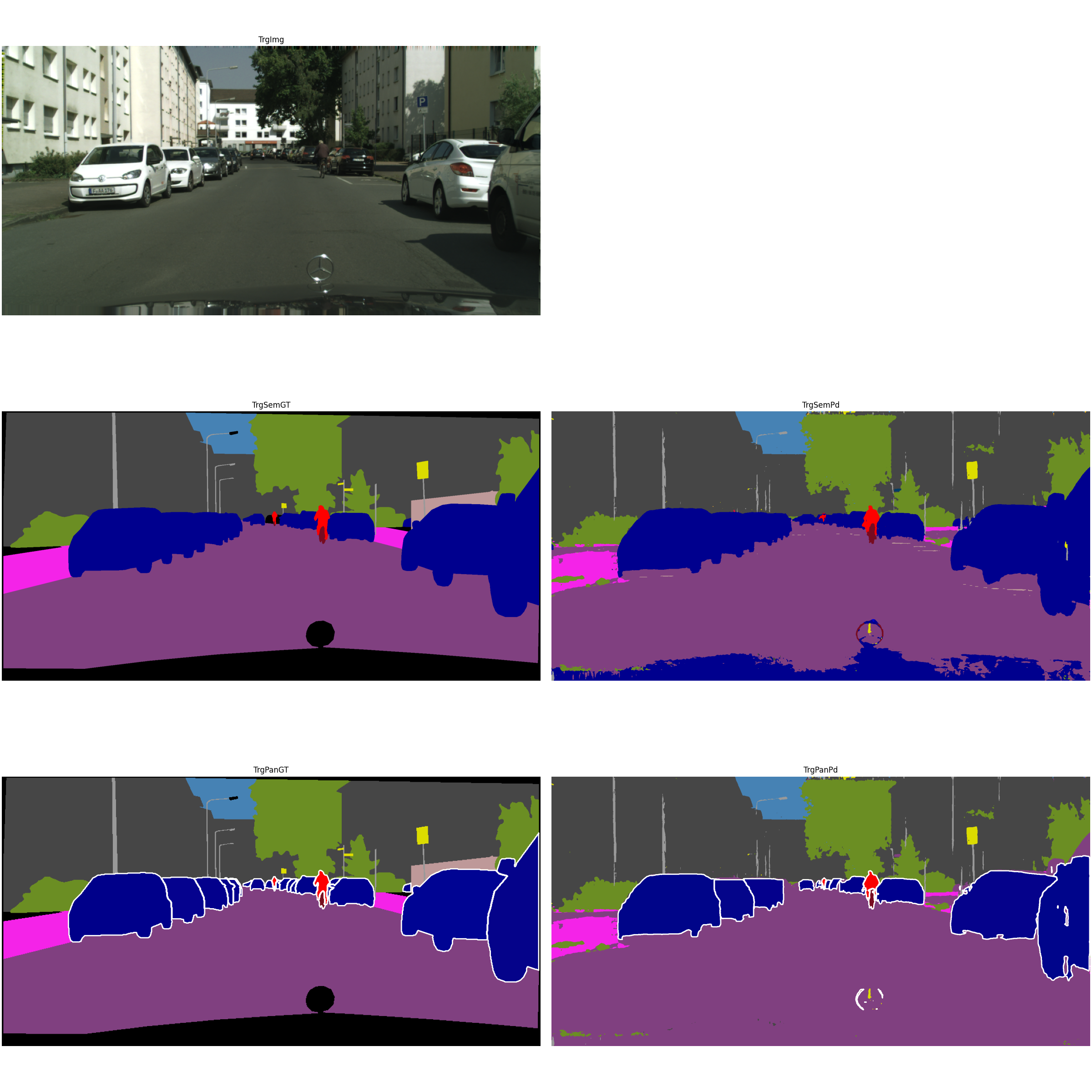}
    \end{subfigure}\hfill
    \begin{subfigure}[b]{0.246\linewidth}
        \includegraphics[trim=10 100 870 1300,clip, width=\linewidth,]{images/frankfurt_000000_000576_leftImg8bit.png}
    \end{subfigure}\hfill
    \begin{subfigure}[b]{0.246\linewidth}
        \includegraphics[trim=870 100 10 1300,clip, width=\linewidth,]{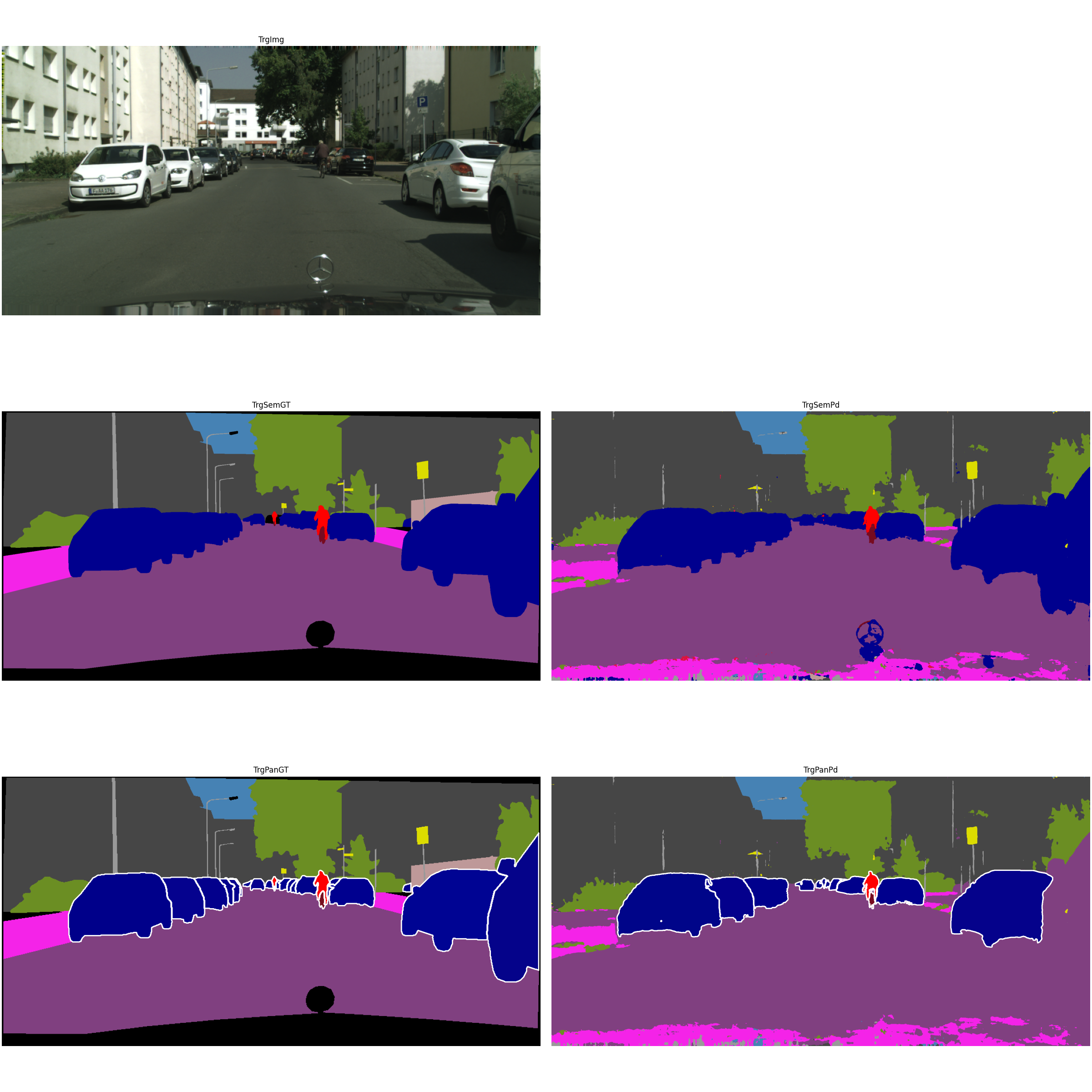}
    \end{subfigure}\hfill
    \begin{subfigure}[b]{0.246\linewidth}
        \includegraphics[trim=870 100 10 1300,clip, width=\linewidth,]{images/frankfurt_000000_000576_leftImg8bit.png}
    \end{subfigure}\\
    \begin{subfigure}[b]{0.246\linewidth}
        \includegraphics[trim=380 1440 1200 195,clip, width=\linewidth,]{images/frankfurt_000000_000576_leftImg8bit.png}
    \end{subfigure}\hfill
    \begin{subfigure}[b]{0.246\linewidth}
        \includegraphics[trim=380 285 1200 1350,clip, width=\linewidth,]{images/frankfurt_000000_000576_leftImg8bit.png}
    \end{subfigure}\hfill
    \begin{subfigure}[b]{0.246\linewidth}
        \includegraphics[trim=1250 285 330 1350,clip, width=\linewidth,]{images/frankfurt_000000_000576_leftImg8bit_reported.png}
    \end{subfigure}\hfill
    \begin{subfigure}[b]{0.246\linewidth}
        \includegraphics[trim=1250 285 330 1350,clip, width=\linewidth,]{images/frankfurt_000000_000576_leftImg8bit.png}
    \end{subfigure}
    \caption{UDA driven by MIC \cite{hoyer2023mic} ($1^{st}$ row) and EDAPS \cite{edaps} ($2^{nd}$ and $3^{rd}$ rows) methods is biased toward more populated (frequent or larger) classes on the source dataset (c), miss-classifying instances of less frequent or smaller classes: false positives examples include instances of \textit{train}, \textit{car} and \textit{person} miss-classified as \textit{truck}, \textit{road} and \textit{car} respectively. (d) GBW improves the classification of under-represented classes.} 
    \label{fig:fig1}
    \vspace{-5mm} %Montalvo, por defecto estaba a -2mm
\end{figure}
\section{Introduction}
%Semantic segmentation is a core task for different applications such as Autonomous driving \cite{HU2022105474}, Medicine \cite{9447006} and Agriculture\cite{MARINO2019337}. In recent years, significant progress has been made in this field, largely attributed to the development of deep neural networks (DNNs) \cite{7298965} and the availability of datasets. However, the progress in dataset quality and quantity has been hindered by the challenges and expenses associated with accurately annotating each pixel \cite{Cordts2016Cityscapes, FERNANDEZMORENO2023106299}.% This concept has undergone numerous enhancements in various aspects, such as expanding the receptive field while maintaining spatial details \cite{7913730}, and incorporating Transformer-based architectures \cite{hoyer2022daformer}.
%only feasible for a small buoyant subset of the research community. For instance,
Dense prediction visual tasks entail a major challenge for the annotation of the required training datasets, as the annotation of a single image for semantic or panoptic segmentation is estimated to take more than an hour for an average human \cite{Cordts2016Cityscapes, FERNANDEZMORENO2023106299}. A bypass solution has been the use of synthetic datasets, as simulated environments can significantly reduce annotation costs. However, a network trained on a synthetic source dataset is expected to underperform when applied to the real target dataset, due to the data distribution co-variate shifts between domains. To mitigate this domain gap, unsupervised domain adaptation (UDA) is defined as the training of a deep learning model focused on generalizing to unlabeled target domains by leveraging labeled data from a source domain \cite{vu2019advent,Tranheden2020DACSDA,hoyer2022daformer,edaps}. 

UDA methods have remarkably progressed in the last few years and  they still are a more effective solution for specific domains than foundational models \cite{chen2023semantic} mainly due to their domain specificity. However, there is still a noticeable performance gap compared to supervised learning \cite{alcovercouso2023soft, Xie2021SegFormerSA,YuanCW20}. One of  the factors contributing to this gap is the complexity of handling class imbalance. As an inherent factor of the data distribution, class imbalance is also subjected to covariant shifts and changes between the source and target domains. Strategies to handle class imbalance can be roughly divided into  \cite{8551020,CIS}: data-level and algorithmic-level techniques. \textbf{Data-level} techniques \cite{hoyer2022daformer,hoyer2022hrda, hoyer2023mic,edaps} can not cope with class-imbalance in dense visual tasks, as they cannot yield uniform class samplings due to the imbalance nature of these tasks, where context is key for prediction and the context of low populated classes is prone to be composed of large populated classes. Therefore, sampling \textit{more} from the low populated classes entails including (eventually more) samples from large populated ones, keeping or even enlarging the imbalance. \textbf{Algorithmic-level} techniques assign different weights to the loss of different classes according to the target class distribution generally under a class-weighting strategy \cite{CIS}. Class-weighing has the effect of differently pondering the contribution to the global training loss of the different classes \cite{8012579}. Class-weighting techniques have been proven ineffective for UDA due to the lack of target domain class-frequencies \cite{8099590, wmmd} and the discredited assumption that target domain class-frequency is equivalent to that of source domain \cite{8099590, wmmd}. Therefore, the only feasible alternative if to estimate the target priors on every model update using pseudo-labels \cite{8099590, wmmd}, inquiring in excessive computational costs, and demotivating the use of algorithmic-level techniques \cite{8012579}.   

UDA methods for image-classification  have reported the advantages of assigning different relevance to the training samples through importance weighting \cite{UDAImportance}. Importance weighting prioritizes data points with high impact on the learning process by focusing on data that generates large gradients during optimization \cite{Katharopoulos2018NotAS}. Adaptive weighting incorporates importance weighting in the training process by giving more importance to certain samples on the fly. This dynamic prioritization changes the learning target, but helps the learning process focusing on examples that have a bigger impact on how the model improves \cite{LOW}. Although per-sample weighting is a key stage in dense visual prediction UDA methods to prevent the adaptation over-fit to fuzzy pseudo-label predictions \cite{Gong_2023_CVPR}, general per-sample adaptive weighting has not been previously explored for dense prediction visual tasks. This is probably due to the memory and computation requirements in estimating a relevance for each pixel.
To address the class-imbalance challenge in UDA, this paper describes a new algorithm-level technique: a gradient-based class-loss weighting (GBW) for UDA that can be applied to dense prediction visual tasks. GBW firstly leverages adaptive weighting for dense visual prediction tasks by considering the gradients of the per-class loss in each iteration as a proxy to the class learning and pondering the learning of the classes such that the overall loss gradient (and hence\textemdash we hypothesize, the learning) is maximized. GBW can be straightforwardly integrated into various UDA methods across different visual recognition tasks, making it highly valuable in practice. The overall effect in UDA of GBW is the alleviation of class imbalance, consistently improving the classification of low populated classes as well as the overall performance. Figure \ref{fig:fig1} shows a visual comparison of the effect of using GBW in state-of-the-art (SOTA) methods HRDA \cite{hoyer2022hrda} for semantic segmentation and EDAPS \cite{edaps} for panoptic segmentation.

\begin{figure}[]
    \centering
    \begin{subfigure}[b]{0.9\linewidth}
    \includegraphics[width=\linewidth]{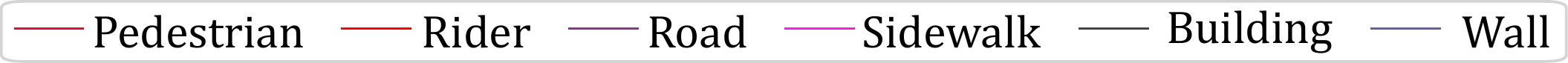}
    \end{subfigure}
    \begin{subfigure}[b]{0.45\linewidth}
    \centering
    \includegraphics[width=.9\linewidth]{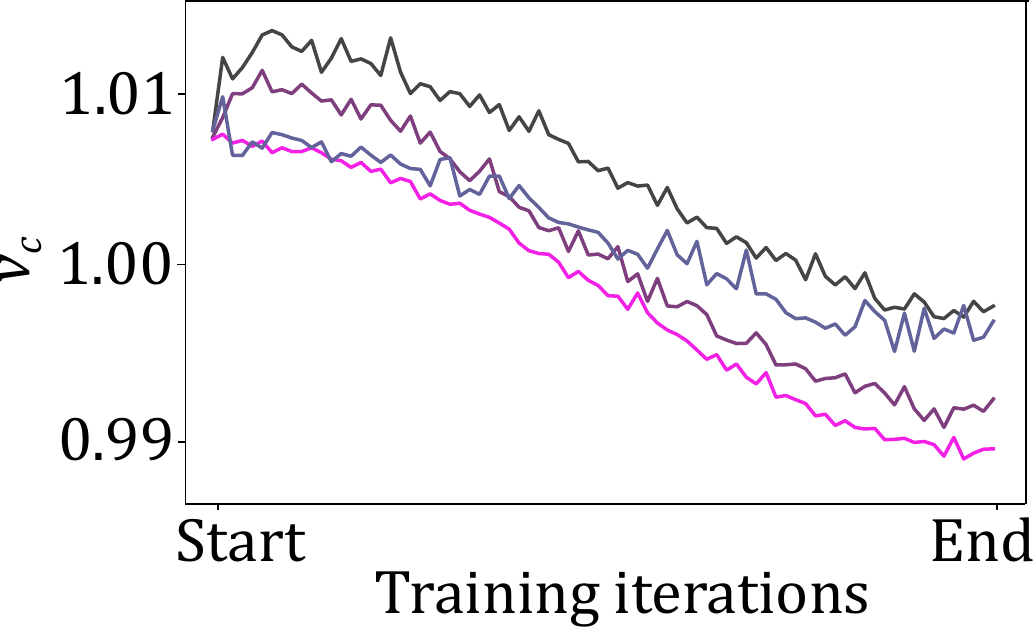}
    \caption{Coarse classes}
    \label{subfig_Coarse}
    \end{subfigure}\hfill
    \begin{subfigure}[b]{0.45\linewidth}
    \centering
    \includegraphics[width=.9\linewidth]{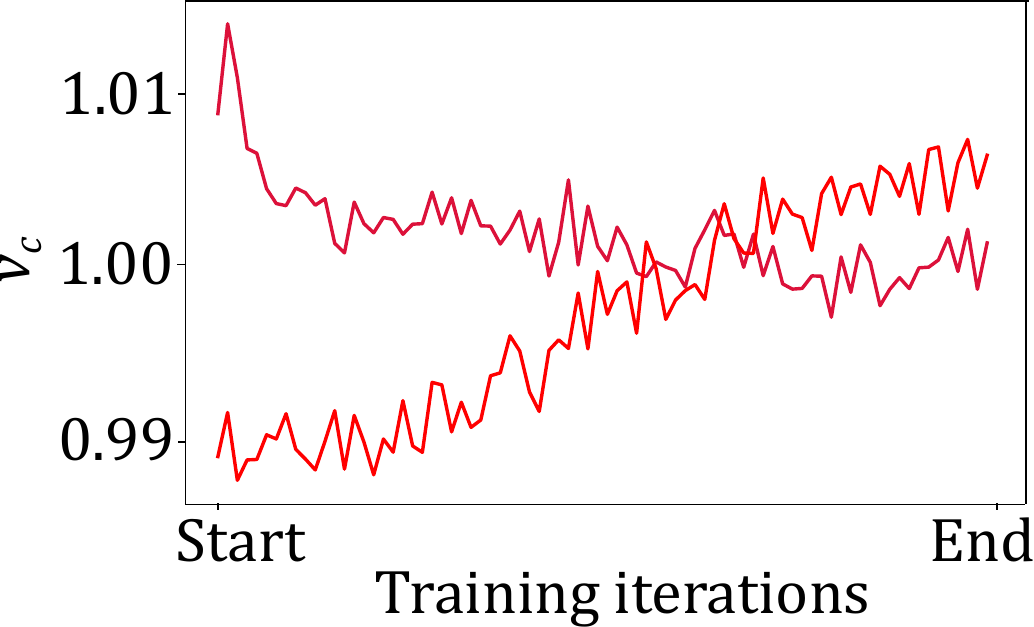}
    \caption{Pedestrian and rider}
    \label{subfig_ped}
    \end{subfigure}\hfill
    
    \caption{Averaged per-class weights ($v_c$) of GBW on the GTA-Cityscapes framework for HRDA method throughout the UDA training process. \ref{subfig_Coarse} shows the evolution for some coarse classes and  \ref{subfig_ped} depicts the complementary evolution for \textit{person} and \textit{rider}.}
    \label{fig:evoweights}
\end{figure}

Leveraging the gradient rather than class frequencies permits the application to both source and target domains, enabling its use for UDA. Performing adaptive weighting at the class rather than at the pixel level, not only effectively enables the use of adaptive weighting techniques for dense visual prediction tasks but also naturally adapts the training to each class learning landscape. 
This enables the learning process to evolve different for each class based on its nature, interactions with other classes, and the training class-wise estimates (see examples of per-class weights evolution in Figure \ref{fig:evoweights}).
% aim to improve the performance and robustness of existing UDA methods. To evaluate the effectiveness of our approach, we conduct extensive experiments using various architectures, methods and datasets. Consistently demonstrating performance gains and emphasizing the complementary nature of our strategy to existing techniques such as sampling.
Experimental evaluation reports significant and consistent performance improvements for different UDA methods (including adversarial training, entropy-minimization, and self-training) on multiple dense prediction visual tasks (semantic and panoptic segmentation), and using different network architectures (CNNs and Vision Transformers). 
%The subsequent sections of this paper are organized as follows: First, we review related work on UDA in semantic segmentation, examining the challenges and existing solutions. Second, we present our proposed per-class weighting approach. Third, we showcase experimental results and discuss the insights derived from our experiments. Finally, we conclude the paper and provide directions for future research, aiming to further enhance UDA methods for semantic segmentation tasks.
\section{Related Work}
%\subsection{Semantic segmentation}
%Building up from DNNs, semantic segmentation architectures have focused on capturing small details while keeping spatial information. To this end, the employment of multiple scale context information can lead to significant performance improvements \cite{hoyer2022daformer, hoyer2022hrda,7913730}. Multi-scale techniques combine features from different scales of an image \cite{hoyer2022daformer,7913730} or rely on a combination of predictions from scaled versions of an image \cite{hoyer2022hrda}. That combination can be implemented either through pooling \cite{7913730} or by employing an attention-weighted fusion  \cite{ hoyer2022daformer, hoyer2022hrda}.
\subsection{Unsupervised Domain Adaptation: UDA}
The goal of UDA is to provide robustness to co-variate shifts on an unlabeled target domain by leveraging training on a labeled source domain. Due to the universality of domain gaps, the use of UDA has been explored in multiple computer vision problems, including image classification \cite{GLC, hoyer2023mic}, semantic segmentation \cite{hoyer2022daformer, hoyer2022hrda} and panoptic segmentation \cite{edaps}. 

%Depending on the number of shared classes, UDA can be classified into: partial-set DA (PDA), where the target label space is included in the source label space; open-set DA (OSDA), where the source label space is included in the target label space; and open-partial-set DA (OPDA), where the target and source label spaces share some classes, but both have unique classes as well.
UDA methods can be classified into three non-exclusive principal categories \cite{10128983}: Input space adaptation \cite{hoyer2022daformer, hoyer2022hrda, hoyer2023mic,Tranheden2020DACSDA,9888149}, feature space adaptation \cite{8578878, vu2019advent} and output space adaptation \cite{AK2023106172,8578878,vu2019advent, hoyer2022daformer}. \textbf{Input space adaptation} methods reduce the domain gap by modifying the style of images following a data transformation protocol based on data augmentations \cite{9888149, Tranheden2020DACSDA}, the mixture of images \cite{Tranheden2020DACSDA} or style transfer \cite{wu2021style, MTAP}. These transformations can be applied to either the source or target domains. \textbf{Feature space adaptation} methods follow the hypothesis that minimizing a discrepancy metric between the source and target model outputs will produce domain-agnostic features. Thereby, a classifier capable of generalizing to both domains indistinguishably can be trained. This alignment is typically driven by adversarial training \cite{8578878, vu2019advent} or by minimizing a suitable metric between features such as L2 distance \cite{road, hoyer2022daformer} or the Maximum Mean Discrepancy (MMD) \cite{10.1145/3474085.3475186, wmmd, 8099590}. \textbf{Output space adaptation} methods rely on the model's predictions during training, often utilizing a process known as self-training, where the model adapts based on its own tentative predictions. This adaptation is typically achieved through the assignment of pseudo-labels, often under some confidence paradigm \cite{hoyer2022daformer, hoyer2022hrda, hoyer2023mic, Tranheden2020DACSDA}. However, as the model is expected to provide false positives, a prominent challenge raises: \textit{concept drifting} \cite{CascanteBonilla2020CurriculumLR, Sammut2010}, which is the event where a model trained on pseudo-labels over-fits to false positives, consequently reducing the performance on the target domain. 

Transversal to all these categories is the problem of class imbalance: Input space adaptation methods employ multiple versions of source images (reinforcing the source class-biases), feature space adaptation aims to generate source agnostic features based on source-biased features, and output space adaptation employ source-biased predictions to generate pseudo-labels. Thus, all UDA stages may benefit from better techniques to handle class imbalance.
%\subsection{Sample-based weighting}
%However, in order to fully exploit their potential on target images they have to be paired with algorithmic techniques \cite{hoyer2022daformer, hoyer2022hrda, hoyer2023mic, Tranheden2020DACSDA} to generate pseudo-labels in the target domain.
% \paragraph{Output space adaptation}
% Output space adaptation approaches can be divided into feature alignment and self-training. Feature alignment approaches follow the hypothesis that minimizing a discrepancy metric between the source and target model outputs will produce domain-agnostic features. So that a classifier capable of generalizing to both domains indistinguishably can be trained.  This metric can be a defined distance such as minimizing the entropy of the predicted target images \cite{vu2019advent} or a trainable one \cite{AK2023106172,8578878}. Trainable metrics rely on auxiliary discriminators to generate domain invariant features in an adversarial manner.

% Self-training approaches generate pseudo-labels for the target domain based on the predictions of the model. As the model will provide false positives, these predictions are filtered by a confidence threshold to avoid whats known as \textit{concept drifting} \cite{CascanteBonilla2020CurriculumLR, Sammut2010}. Concept drifting is known as the event where a model trained on pseudo-labels over-fits to false positives, thus, reducing at each time-step the performance on the target domain. This threshold can be manually defined \cite{Tranheden2020DACSDA} or dynamically defined \cite{9710267,zheng2020unsupervised,zheng2021rectifying}.

\subsection{Handling class imbalance in UDA}

In real-world scenarios, accurately classifying examples from an infrequent class poses a significant challenge known as the class imbalance problem. For image classification, competitive datasets, such as CIFAR-10/100 \cite{CIFAR}, ImageNet \cite{5206848}, Caltech-101/256\cite{griffin_holub_perona_2022}, and MIT-67 \cite{quattoni2009recognizing}, deliberately hampers class imbalance by ensuring all classes have a minimum representation \cite{8012579}. 
Unfortunately, this balanced-samples design becomes infeasible in the context of dense tasks such as semantic segmentation. Certain classes, regardless of the number of available images, inevitably exhibit significantly smaller representations compared to broader classes. Furthermore, for some domains it is not possible to obtain more images from certain classes, such as rare illnesses for medical tasks. Therefore, mitigating the adverse effects of imbalanced class distributions emerges as a transversal research problem \cite{8012579,  10.1093/bioinformatics/btm158, 10.1007/978-3-319-67389-9_44}. 

Techniques to handle class imbalance in machine learning can be broadly classified into two main categories \cite{8551020,CIS}: data-level and algorithmic-level techniques. Data-level techniques try to balance the data by reducing the likelihood of selecting  majority class samples for training (undersampling) or increasing the likelihood of minority class sampling (oversampling)\cite{8551020}. On the other hand, algorithmic-level techniques modify the learning process to tackle the bias produced by the imbalanced data. These techniques are typically implemented with a weight schema assigning penalties to each class \cite{8012579,CIS,classweights,Lin_2017_ICCV}. Alternatively, hard class-weighting strategies have been proposed for semantic segmentation so that all classes contribute equally in the loss \cite{10.1007/978-3-319-67558-9_28,YEUNG2022102026,9338261}. However, these strategies are not well suited for densely populated images, as result in small objects accounting equally in the loss to classes composing most of the scene.%  Increasing the cost of the minority group is equivalent to increasing its importance, thus, decreasing the likelihood that the model will incorrectly classify instances from this group \cite{classweights, 8012579}.

In the context of UDA for dense tasks, data-level techniques are the default and typically the only imbalance techniques employed \cite{hoyer2022hrda,hoyer2022daformer,hoyer2023mic,edaps}.  This is mainly because standard data-level techniques rely on the source domain class frequency and algorithmic-based techniques are not effective without target domain class-frequency \cite{8099590, wmmd}. Although a few works have explored algorithmic-level techniques, these either require the estimation of the target dataset class-frequency at every model update \cite{8099590, wmmd} or are focused on relaxing the filtering criteria of pseudo-labels for low-populated classes\cite{Zou_2018_ECCV, saporta2020esl}. The former becomes unpractical for dense tasks, as such estimation must be performed at every model update making the process very time consuming. The latter does not tackle the root problem because, as the model is trained with the source biases, the pseudo-labels are more prone to classify least represented classes as other densely populated classes in the source set. Altogether, current state-of-the-art UDA methods for dense tasks rely solely on the employment of sampling techniques \cite{hoyer2022hrda,hoyer2022daformer,hoyer2023mic,edaps}, discarding the possible benefits of joint training with data and algorithmic imbalance techniques \cite{8551020, CIS}. 

This paper describes a new algorithmic-level technique complementary with data-level techniques that can be applied to source and target domain, not requiring target domain information, by firstly tailoring adaptive weighting \cite{LOW} to a class-wise paradigm and incorporating it into the UDA learning process.

\section{Method}

\subsection{The learning of dense prediction visual tasks}
Let $\hat{\mathbf{y}}_{i,t} = G(x_i;\theta_t)$ be the output of a deep learning model parameterized by $\theta$, at the $t^{th}$ training step, given an input sample $\mathbf{x}_i$. In the context of classification, $\hat{\mathbf{y}}_{i,t}$ is typically a per-class probability vector. Note that for dense tasks such as semantic and panoptic segmentation, a sample is considered to be a pixel from the image. %Let each pixel of the sample image $i$ be $x_i^{h,w}$ for  $(h,w)\in [1,H] \times [1,W]$ where $H,W$ are the height and width of the image. Let $y_i \in [1,C]^{[H,W]}$ be the corresponding segmentation map, where  $C$ is the number of semantic categories.

The learning is typically driven by maximizing the probability of the expected class $y_i$. This can be segregated on a per-class basis: 
% \begin{equation}
% \begin{split}
%     &L(\hat{\mathbf{y}}_{i,t},y_i)  =\\
%     &{\begin{bmatrix}{\mathbbm{1}_{y_i=1}l(\hat{\mathbf{y}}_{i,t}, 1)&\hdots&\mathbbm{1}_{y_i=C}l(\hat{\mathbf{y}}_{i,t}, C)}\end{bmatrix}},
% \end{split}
% \end{equation}

\begin{equation}
    \mathcal{L}(\hat{\mathbf{y}}_{i,t},y_i)  = \sum_{c=1}^{C} \frac{l_c(\hat{\mathbf{y}}_{i,t},y_i)}{|y_i = c|},
\end{equation}
where $C$ is the number of classes, $L(\hat{\mathbf{y}}_{i,t},y_i)$ is the aggregated loss, $l_c$ the probability distance of the prediction to a given class and $|\cdot|$ the cardinal function. For source domain labels and target domain pseudo-labels, $l_c$ is typically the cross-entropy loss: $l_c(\hat{\mathbf{y}}_{i,t}, y_i) = -\delta[c-y_i]log(\hat{y}_{i,t})$. Furthermore, for the target domain some regimes also minimize the entropy of the predictions\cite{vu2019advent}: $l_c(\hat{\mathbf{y}}_{i,t}) = -\hat{y}_{i,t}^clog(\hat{y}_{i,t}^c)$.

Including into the formulation a set of per-class weights $\mathbf{v} \in \mathbb{R}^C$, the loss for each sample is defined by:

\begin{equation}
   \mathcal{L}(\hat{\mathbf{y}}_{i,t},y_i;\mathbf{v})  = \sum_{c=1}^{C}v_c\frac{l_c(\hat{\mathbf{y}}_{i,t},y_i)}{|y_i = c|}.
\end{equation}
% \begin{equation}
%     \mathcal{L}(\hat{\mathbf{y}}_{i,t},y_i;\mathbf{v})  = \sum_{c=1}^{C} v_c\delta[n-c](\hat{\mathbf{y}}_{i,t},y_i).
% \end{equation}
When no class weighting is applied $\mathbf{v}$ is a vector of one value elements of size $C$.

The learning  of $\theta$ is driven by the minimization of the loss $\mathcal{L}$ over $N$ samples:
\begin{equation}
    \theta^* = arg \min_\theta \frac{1}{N}\sum_{i=1}^N \mathcal{L}(\hat{\mathbf{y}}_{i,t}, y_i).
\end{equation}

This optimization is often solved through stochastic gradient descent (SGD) with mini-batches, leading to the update equation:
\begin{equation}
    \theta_{t+1} = \theta_t - \eta \frac{1}{M} \sum_{i=1}^M \nabla_{\theta_t} \mathcal{L}(\hat{\mathbf{y}}_{i,t}, y_i; \mathbf{v}_t),
    \label{eq:optW}
\end{equation}
where $\eta$ represents the learning rate, $t$ the optimization step, and the samples $i=1,...,M$ are selected randomly from the training set.

% \paragraph{SGD with weights}
% In the update, all samples and classes contribute equally to the optimization of the network parameters. However, weighting strategies enforce the network to pay attention to more relevant samples or classes by assigning them higher weights. These weights can be sample-based, denoted by $\mathbf{p}_i^t \in \mathbb{R}_+^{H,W}$, or class-based, denoted by $\mathbf{v}_t \in \mathbb{R}_+^C$.

%In the update, all samples and classes contribute equally to the optimization of the network parameters. However, weighting strategies enforce the network to pay attention to more relevant samples or classes, by assigning them higher weights. These weights can be sample based: $p_j^t\in \mathbb{R}^+$ or class based: $v_c^t\in \mathbb{R}^+| c \in [1,C]$. These weights are incorporated in the update equation by multiplying the sample loss by a specific weight:
% \begin{equation}
%     \theta_{t+1} = \theta_t - \eta \frac{1}{M} \sum_{i=1}^M \mathbf{p}_i^t \nabla_{\theta_t} \mathcal{L}(\hat{y}_{i,t}, y_i; \mathbf{v}_t).
%     \label{eq:optW}
% \end{equation}

\subsection{Gradient based weighting (GBW)}
\begin{figure*}[tp]
    \centering
    \includegraphics[width=.8\linewidth]{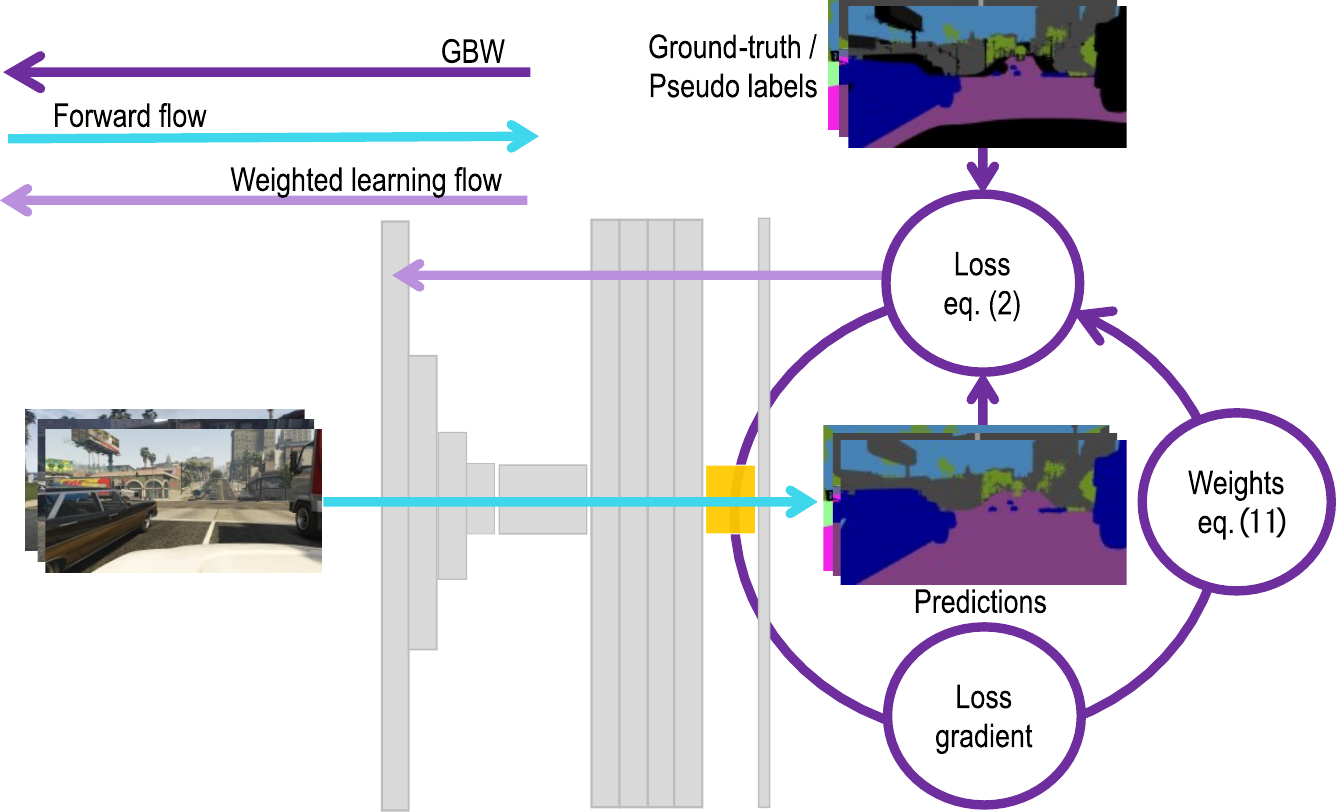}
    \caption{Semantic segmentation using GBW. Given a forward pass of an image and the respective per-class loss ($ l_c$). First, the gradients ($||\nabla_{\theta_t} \overline{l_c}||^2$) are estimated with the gradient of the last layer (shadowed in orange) to compute the per-class weights $\mathbf{v}$. Second, the backward pass is performed \textit{wrt} the weighted cross entropy $\mathcal{L}(\hat{\mathbf{y}}_{i,t},y_i;\mathbf{v}_t)$.}
    \label{fig:process}
    \vspace{-5mm}
\end{figure*}
We adapt \cite{LOW} by following the assumption that minimizing a class loss should make the model more prone to classify that class, therefore, by dynamically setting optimal weights to decrease the overall gradient, the model should avoid over-fitting to broad classes. Additionally, this avoids manually setting class weights which could lead to over-classifying small classes. To that end, GBW estimates class weights at each learning step given the average per-class loss ($\overline{l_c}$) gradients:
\begin{equation}
    \mathbf{v}^* = arg \min_\mathbf{v} \overline{l_c}(\hat{\mathbf{y}}_{i,(t+1)}, y_i;\mathbf{v}_t) - \overline{l_c}(\hat{\mathbf{y}}_{i,t}, y_i;\mathbf{v}_t).
    \label{eq:difference}
\end{equation}
Using linear approximation, one can approximate the term $\overline{l_c}(\hat{\mathbf{y}}_{i,(t+1)}, y_i;\mathbf{v}_t)$:
\begin{equation}
\overline{l_c}(\hat{\mathbf{y}}_{t+1}, y_i;\mathbf{v}_t) \leq  \lbrace \overline{l_c}(\hat{\mathbf{y}}_{i,t}, y_i;\mathbf{v}_t) +  (\nabla_{\theta_t}\overline{l_c}(\hat{\mathbf{y}}_{i,t}, y_i;\mathbf{v}_t))^T(\theta_{t+1}-\theta_{t})\rbrace
\label{eq:taylor}    
\end{equation}
Note that $\nabla_{\theta_t} \overline{l_c}(\cdot;\mathbf{v}_t) = \frac{1}{C}\sum_{c=1}^{C}v_c\nabla_{\theta_t}l_c$, making the operation linearly dependant on the per-class losses.
\begin{equation}
    \begin{split}
        &\overline{l_c}(\hat{\mathbf{y}}_{i,t}, y_i;\mathbf{v}_t)- \overline{l_c}(\hat{\mathbf{y}}_{i,(t+1)}, y_i;\mathbf{v}_t) \\
        &\text{including Eq \ref{eq:taylor}}\\
        &\begin{split}
            \leq 
            \overline{l_c}(\hat{\mathbf{y}}_{i,t}, y_i;\mathbf{v}_t)& -\overline{l_c}(\hat{\mathbf{y}}_{i,t}, y_i;\mathbf{v}_t)
             - (\nabla_{\theta_t} \overline{l_c}(\hat{\mathbf{y}}_{i,t}, y_i)^T(\theta_{t+1}-\theta_{t})  
        \end{split}\\
        &= - (\nabla_{\theta_t} \overline{l_c}(\hat{\mathbf{y}}_{i,t}, y_i;\mathbf{v}_t)^T(\theta_{t+1}-\theta_{t}) \\
        &\text{including Eq \ref{eq:optW} and simplifying}\\
        %&= - (\nabla_{\theta_t} l(G(x_i;\theta_t), y_i;\mathbf{v}_t)^T( \theta_t - \eta p_i^t\nabla_{\theta_t} l(G(x_i;\theta_t), y_i; \mathbf{v}_t)-\theta_{t}) \\
        &= (\nabla_{\theta_t} \overline{l_c}(\hat{\mathbf{y}}_{i,t}, y_i; \mathbf{v}_t)^T(\eta \nabla_{\theta_t} \overline{l_c}(\hat{\mathbf{y}}_{i,t}, y_i; \mathbf{v}_t))\\
        &= \eta ||\nabla_{\theta_t} \overline{l_c}(\hat{\mathbf{y}}_{i,t}, y_i;\mathbf{v}_t)||^2.
    \end{split}
\end{equation}

Class weights can be calculated through the following optimization problem:
\begin{equation}
\begin{split}
    \mathbf{v}^* = arg \min_{\mathbf{v}} - ||\nabla_{\theta_t}\overline{l_c}(\hat{\mathbf{y}}_{i,t}, y_i;\mathbf{v}_t)||^2\\
    \text{subject to } v_c \geq 0, \forall c \in [1,C]\\
    %\sum_{c=1}^C v_c =C,\\
\end{split}    
\end{equation}
This optimization can be efficiently computed by expressing the problem as a quadratic programming problem \cite{Vavasis2009}:

\begin{equation}
\begin{split}
    \text{minimize }\textbf{v}_t^TQ\textbf{v}_t-(||\nabla_{\theta_t}\overline{l_c}||^2)^T\textbf{v}_t\\
   \text{subject to } -I\textbf{v}_t\leq \textbf{0},
\end{split}
\end{equation}  
where $\textbf{0}$ is a vector filled with zeros and $I$ is the identity matrix. The only problem left is defining the $Q$ matrix. Through initial experimentation, we found that this optimization would lead to highly variable weights which disrupt learning. Therefore, we opt to define $Q$ as a regularization matrix: $Q=\lambda I$, where $\lambda$ is a regularization parameter. Specifically, we  impose an $L_2$ regularization on $\textbf{v}_t$:

\begin{equation}
\begin{split}
    \mathbf{v}^* = arg \min_{\mathbf{v}} -(||\nabla_{\theta_t}\overline{l_c}||^2)^T\textbf{v}_t + \lambda||\textbf{v}_t||^2\\
   \text{subject to }  v_c \geq 0, \forall c \in [1,C]\\
\end{split}
\end{equation} 

Additionally, to maintain a similar scale to the original loss, we impose an additional constraint that ensures the weights sum up to the number of classes. 

\begin{equation}
\begin{split}
    \mathbf{v}^* = arg \min_{\mathbf{v}} -(||\nabla_{\theta_t}\overline{l_c}||^2)^T\textbf{v}_t + \lambda||\textbf{v}_t||^2\\
   \text{subject to }  v_c \geq 0, \forall c \in [1,C]\\ \text{ and }
   \sum_{c=1}^C v_c =C,
\end{split}
\label{eq:lambda}
\end{equation} 
the last constraint also promotes that all batches are weighted similarly in the loss, as without it, we seldom noticed large discrepancies for some instances and training iterations that were precluding a smooth learning process.

Overall, the complexity of GBW scales with the number of classes:  $O(C)^2$, making its resolution feasible for large datasets. However, to further facilitate the computation of the gradient norms without performing back-propagation throughout the entire model, the gradient norms are further approximated by calculating the gradient of the loss function with respect to the pre-activation outputs of the last layer in the model. The use of these pre-activation gradients, as demonstrated in previous studies \cite{LOW, Katharopoulos2018NotAS}, significantly simplifies the weight computation process and substantially reduces computational complexity, while preserving a significant correlation to the true gradients. Figure \ref{fig:process} provides a visual summary of our GBW methodology for semantic segmentation.

%\subsection{Generalization of the per-class-weights}
\paragraph{Combination with previous per-sample weighting}
In UDA, pseudo-labels are commonly employed to reinforce the classification performance on the target data. These pseudo-labels usually are weighted at a sample level to avoid over-fitting to fuzzy predictions \cite{hoyer2022daformer,hoyer2022hrda, hoyer2023mic, Tranheden2020DACSDA, Gong_2023_CVPR, WANG2023103743, Shen_2023_CVPR, Li_2023_CVPR}. Notably, neglecting these pseudo-labels in the optimization goal leads to the class-weights counteracting the confidence weighting in the loss. To prevent this, we modify the optimization goal in Equation \ref{eq:lambda} by incorporating the pseudo-label confidence $p_i$:  
\begin{equation}
 \mathbf{v}^* = arg \min_{\mathbf{v}} - (||\nabla_{\theta_t}\sum_{i=1}^{M}p_i\cdot l_c(\hat{\mathbf{y}}_{i,t}, y_i)||^2)^T + \lambda||\textbf{v}_t||^2 .
\end{equation}

%\paragraph{Extrapolation to unsupervised losses}
%This weighting can be easily extrapolated to any clustering or entropy-based loss. As long as the loss can be segregated into groups, the average loss for each group can be computed. Therefore, the average per-group gradient can be also computed by changing the loss function in Eq \ref{eq:terms}.

\section{Experimental Exploration}
%In this section, we provide a comprehensive experimental analysis to evaluate the effectiveness of our proposed weight merging technique for Unsupervised Domain Adaptation (UDA). We begin by describing the experimental setup, including the datasets used, evaluation metrics employed, and training parameters applied. Subsequently, we conduct an ablation study to determine the optimal value of the weighting parameter $\lambda$, which remains consistent throughout the rest of the paper. Next, we investigate the impact of our per-class weights on various training approaches and architectures. We analyze their performance and assess their ability to improve adaptation across different domains. Additionally, we compare our weight merging technique with other existing weighting proposals for UDA.%, aiming to highlight its advantages and provide a comprehensive understanding of its efficacy.
\subsection{Setup}
\paragraph{Semantic Segmentation}

We explore the benefits of the proposed GBW method on two popular UDA scenarios defined by the adaptation for semantic segmentation of the source datasets GTA\cite{Richter_2016_ECCV} and Synthia \cite{Ros2016} to the target dataset Cityscapes \cite{Cordts2016Cityscapes}.  \textbf{GTA} is a synthetic dataset comprising 25K images and sharing 19 semantic classes with Cityscapes. \textbf{Synthia} is a synthetic urban scenes dataset for semantic segmentation, composed of 9.5K images  and with 16 common semantic classes with Cityscapes. \textbf{Cityscapes} is a real-image dataset with urban scenes generated by filming with a camera inside of a car while driving through different German cities. It consists of 3K images for training and 0.5K images for validation. For evaluation, we rely on the common metric used for semantic segmentation: the per-class intersection over union (IoU) \cite{Everingham10thepascal}, between the model prediction and the ground-truth label. IoU measures at pixel-level the relationship between True Positives (TP), False Positives (FP) and False Negatives (FN): $IoU = \frac{TP}{TP+FP+FN}$.
\paragraph{Panoptic Segmentation}
Similarly to Semantic segmentation, we measure the benefits of GBW for UDA between the synthetic source dataset Synthia \cite{Ros2016} and the real target dataset Cityscapes \cite{Cordts2016Cityscapes}. These are evaluated through common Panoptic segmentation metrics: mean Segmentation Quality (mSQ), mean Recognition Quality (mRQ) and mean Panoptic Quality (mPQ) \cite{Kirillov_2019_CVPR}. The mSQ measures the closeness of the predicted segments with their ground truths, mRQ is equivalent to the F1 score and the mPQ presents a global quality of the panoptic segmentation by combining at a per-class level the SQ and RQ: PQ = SQ$\times$RQ. Note that improving the recall of an algorithm mainly affects mRQ, as mSQ is mostly driven by the segmentation quality of the TP \cite{Kirillov_2019_CVPR}.

\paragraph{Training parameters} The architectures employed and all training parameters but our proposed weight regularization in equation \ref{eq:lambda}, are as in the original papers \cite{hoyer2022daformer,hoyer2022hrda,hoyer2023mic,Tranheden2020DACSDA,8578878,vu2019advent, edaps}.

\subsection{GBW for UDA in Semantic Segmentation}
\begin{table}[]
    \centering
    \resizebox{\linewidth}{!}{%
    
    \begin{tabular}{l l| c |cc |cc}
         %Architecture
          Architecture & UDA Method & data-level &\multicolumn{2}{c}{GTA}&\multicolumn{2}{c}{Synthia}\\
         & & & w/o GBW& w/ GBW & w/o GBW& w/ GBW  \\\toprule
         Convolutional & Output \cite{vu2019advent} & $\times$&43.8&46.1 &38.1& 41.2 \\
         & Feature \cite{8578878} & $\times$ &42.4 & 45.5& 46.7* & 50.7*\\
         & Input \cite{Tranheden2020DACSDA} & $\times$ &52.1 & 55.4 & 48.3 &50.1\\\midrule
         %DAFormer& DACS\cite{Tranheden2020DACSDA} & $\times$&61.7\\
         %& Ours-DACS & $\times$&65.2 & 3.5\\
         Transformer& Output + Input\cite{hoyer2022daformer} & $\checkmark$&68.3&69.2 & 60.9&62.3\\
         &  Output + Input\cite{hoyer2022hrda}& $\checkmark$&73.8&74.7& 65.8&66.8\\
         &  Output + Input\cite{hoyer2023mic}& $\checkmark$&75.9&76.4& 67.5& 68.8\\ \bottomrule
    \end{tabular}%
    }
    \caption{Comparison of UDA methods for the GTA-Cityscapes and Synthia-Cityscapes semantic segmentation frameworks for different architectures and methods.}
    \label{tab:comparison}
    \vspace{-10mm}
\end{table}

\paragraph{Incorporation of GBW into UDA methods}
 Table \ref{tab:comparison} measures the performance gain of key state-of-the-art UDA methods in the GTA-Cityscapes and Synthia-Cityscapes scenarios when GBW is incorporated. Methods are arranged according to the use (marked with a \checkmark) or not (marked with a $\times$) of some form of data-level class imbalance techniques. The results show that incorporating our per-class weighting strategy consistently leads to performance gains across various architectures and methods. Notably, the largest gains are observed for methods that do not employ data-level class imbalance techniques, with improvements of up to 3.3 points in mIoU for GTA-Cityscapes and 4.0 for Synthia-Cityscapes. BW provides moderate gains also for methods that already utilize data-level class imbalance techniques, underlining its complementary nature to these strategies.

\begin{table}[]
    \centering
    \setlength{\tabcolsep}{15pt}
    \resizebox{\linewidth}{!}{%
        \begin{tabular}{l l c c}
         %Architecture
         Architecture & Method & GTA-Cityscapes  & Synthia-Cityscapes\\\toprule
         Convolutional%& AdaptSegNet\cite{8578878} &42.4&46.7 \\
         %&Advent \cite{vu2019advent}  &43.8&38.1\\
         & CRA\cite{WANG2023103743} & 46.7&49.3\\
         & Zhang et al. \cite{zhang2023black} & 48.2&48.4\\
         %& DACS\cite{Tranheden2020DACSDA}&52.1&48.3  \\
         & GBW (DACS) & \textbf{55.4} &\textbf{50.1}\\\midrule
        
        %& CRA\cite{WANG2023103743} & 43.4\\
        %& Ours & \textbf{45.5}\\\hdashline
        %
        %&Ours & 50.1\\
        Transformer%&  CRA\cite{WANG2023103743} & 55.5&57.2\\
         %& FAFS\cite{9888149} & 56.3&\\
         %&CONFETI\cite{Li_2023_CVPR}&63.3\\
         %&DAFormer\cite{hoyer2022daformer}&68.3&60.9\\
         &ExpCons\cite{expCons}& 69.6&61.5\\
         %& Ours & \textbf{68.7} \\\hdashline
         & DiGA\cite{Shen_2023_CVPR}& 70.0&62.1\\
         %& FREDOM\cite{Truong_2023_CVPR}& 73.6\\
         %&HRDA\cite{hoyer2022hrda}&73.8&65.8\\
         %
         %& F-RMM\cite{Gong_2023_CVPR}& 74.4\\
         &MIC\cite{hoyer2023mic}&75.9&67.5\\
         & GBW (MIC)& \textbf{76.4}&\textbf{68.8}\\\bottomrule
    \end{tabular}}%
    \caption{Comparison of leading GTA-Cityscapes and Synthia-Cityscapes UDA semantic segmentation frameworks for convolutional- and transformer-based architectures.}
    \label{tab:weightingGTA}
    \vspace{-10mm}
\end{table}

\paragraph{Comparison with state-of-the-art} Table \ref{tab:weightingGTA} provides a comprehensive comparison of the currently state-of-the-art convolutional-based and transformer-based models. Notably, the inclusion of GBW in the learning of MIC \cite{hoyer2023mic} results in performance gains compared to the other analysed state-of-the-art methods for both scenarios. Importantly, GBW achieves these results without the need for computationally expensive transformations such as style transfer \cite{WANG2023103743}, or feature prototyping \cite{9888149,zhang2023black}.%, which require memory banks or additional models. 

\subsection{GBW for UDA in Panoptic segmentation}
For panoptic segmentation UDA, we explore the benefits of including GBW in the learning process of the state-of-the-art method EDAPS \cite{edaps}. In the Synthia-Cityscapes scenario (Tab. \ref{tab:Panoptic}), the use of GBW provides improvements across all metrics. Specifically of +0.5 mSQ, +1.8 mPQ and +2.3 mRQ. As mSQ measures the segmentation quality of the true positives, it is mostly driven by the segmentation capabilities of the network, thus, the gain in terms of mSQ is narrower. To the best of our knowledge, this GBW-enhanced EDAPS setup is the current state-of-the-art for UDA in panoptic segmentation across all metrics.
\begin{table}[t]

    \centering
    \setlength{\tabcolsep}{27pt}
    \resizebox{\linewidth}{!}{%
    \begin{tabular}{l c c c c}
         Method& mPQ & mSQ & mRQ\\\toprule
         %AdvEnt \cite{vu2019advent}& 28.1&65.6&36.3\\
         CVRN \cite{Huang_2021_CVPR}& 32.1&66.6&40.9\\
         UniDAPS \cite{10203383}& 33.0&64.7&42.2\\
         EDAPS \cite{edaps}& 41.2&72.7&53.6\\\hdashline
         GBW (EDAPS)&\textbf{43.0}&\textbf{73.2}&\textbf{55.9}\\\bottomrule
    \end{tabular}}
    \caption{Performance comparison of state-of-the-art Synthia-Cityscapes UDA panoptic segmentation frameworks. Best results indicated in bold. }
    \label{tab:Panoptic}
    \vspace{-10mm}
\end{table}
\subsection{Analysis of GBW}
\paragraph{Combination with data-level class imbalance techniques}

Currently, the default technique for handling class imbalance in UDA is rare class sampling \cite{hoyer2022daformer}. Therefore, we propose to analyze the impact of GBW when used in conjunction and separated from the standard data-level technique. Table \ref{tab:sampling} presents a comparison of various configurations incorporating per-class weights and class-uniform sampling. The results emphasize the substantial impact of these techniques in the performance of the DAFormer method for semantic segmentation UDA \cite{hoyer2022daformer}: the combined approach yields +7.5 points increase in terms of mIoU.  %These findings underscore the importance of carefully considering and integrating both weighting and sampling techniques when tackling unbalanced classification challenges in semantic segmentation.% By employing these strategies, researchers and practitioners can develop more effective and robust models for a wide range of applications in computer vision and beyond.

\begin{table}[h]
\vspace{-3mm}
    \centering
    \begin{subtable}{.48\linewidth}
    %\resizebox{\linewidth}{!}{%
        \begin{tabular}{l| c|cc |c}
         %Architecture
         & None &w/GBW&w/S&w/GBW+S\\\toprule
         mIoU&61.7&65.2&68.3&69.2\\
         Gain&0&3.5&6.6&7.5\\\bottomrule
         % GBW & Sampling & mIoU & Gain\\\toprule
         % $\times$ & $\times$ & 61.7& \\
         % $\checkmark$ & $\times$ & 65.2& 3.5\\
         % $\times$ & $\checkmark$ & 68.3& 6.6\\
         % $\checkmark$ & $\checkmark$ & 69.2& 7.5\\\bottomrule
    \end{tabular}%}
    \caption{Data-level techniques on DAFormer \cite{hoyer2022daformer}.}
    \label{tab:sampling}
    \end{subtable}\hfill
    \begin{subtable}{.49\linewidth}
    \resizebox{\linewidth}{!}{%
        \begin{tabular}{lccccc}
        Loss & wo& w/IF & w/PF &w/LBW &w/GBW\\\midrule
        CE & 73.8 & 71.3 & 30.7&72.9& 74.7\\
        %Dice-loss \cite{10.1007/978-3-319-67558-9_28}& 55.4 & 51.8 & 7.31 & 54.7\\
        FL \cite{Lin_2017_ICCV}& 68.9 & 68.5 & 64.0 &68.9& 70.1\\\bottomrule
    \end{tabular}}
    \caption{Algorithmic-level techniques on HRDA \cite{hoyer2022hrda}.}
    \label{tab:losses}
    \end{subtable}
    \caption{Performance comparison with alternative data and algorithmic level class imbalance techniques on the GTA-Cityscapes semantic segmentation framework \cite{hoyer2022daformer}. Key.S: Sampling. CE: Cross-entropy. FL: Focal-loss. IF: image frequency. PF: pixel frequency. LBW: loss-based weighting.}
    \label{tab:imbalance-tech}

    \vspace{-13mm}
\end{table}

\paragraph{Effect on different segmentation losses}
Table \ref{tab:losses} compiles the performances of incorporating in the learning process state-of-the-art class weighting strategies, including regular class weights \cite{frequency}: the inverse frequency on the source set, either considering the pixel-level frequency (PF) \cite{alonso2021semi} or the image-level frequency (IF) \cite{Chen_2019_ICCV}, and a class weighting based on the per class loss rather than their gradient (LBW) \cite{9897273}. Reported performance corroborate the previous evidence that weighting the loss based on source domain priors do not transfer well to the target domain due to the distribution shift. Regarding the frequency source, PF results in drastically higher weights for underrepresented classes compared to IF, turning the model prone to classify broad structures as low-represented categories, thus presenting a significant drop in performance. Meanwhile, IF weighting assigns more uniform weights, thus, presenting better performance than PF, but dropping the performance of broad categories compared to GBW. The use of LBW results in second performance, but suffers from a stable tendency that prevents it from quickly adapting to the learning outcomes as GBW. %Additionally, alternative losses such as Dice-loss its not suited for really small details which may be not segmented due to the capabilities of the network and carries most of the loss. Therefore, dampening the learning of the model.

\paragraph{Per-class improvements}
%Figure \ref{fig:per-class} provides a per-class comparison between the state-of-the-art HRDA model \cite{hoyer2022hrda} and our results when using the HRDA framework with our per-class weights. Although our model shows marginal improvements for most classes, it is observed that the ``sidewalk",  ``motorcycle" and ``rider" classes do not benefit from our approach.

\begin{figure}[]
    \centering
    \vspace{-3mm}
    \begin{subfigure}[b]{0.48\linewidth}
        \includegraphics[trim=10 10 0 0,clip, width=1.1\linewidth,]{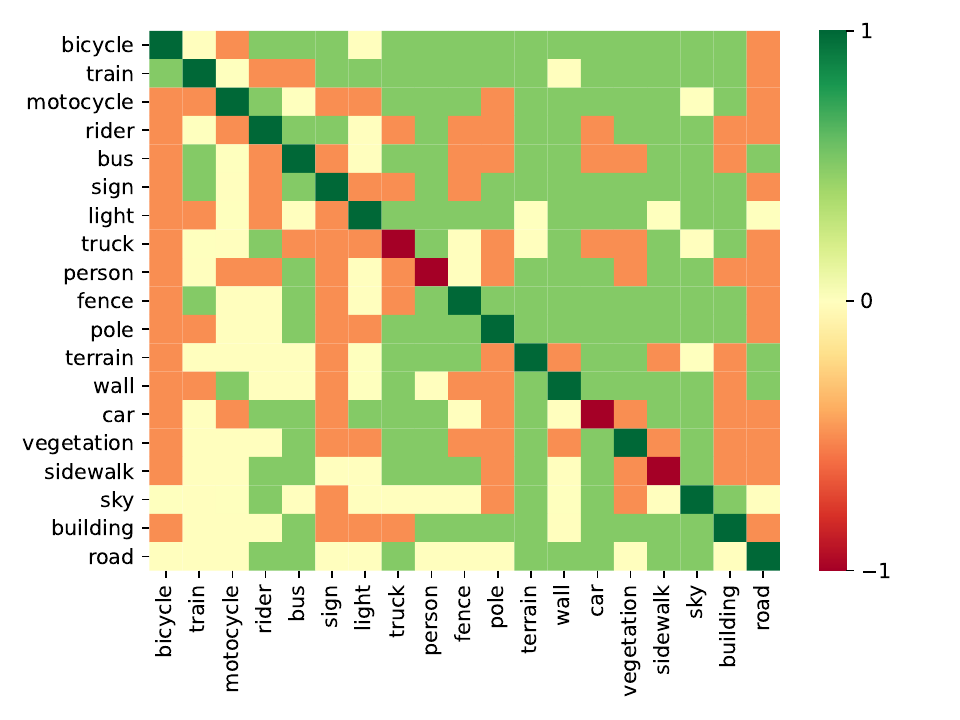} % te he hecho la figura un poco más grande, solo el gráfico Montalvo
        \caption{Relative discrepancy matrix (green/red indicates a rise/drop in performance, yellow indicates equal performance) between our GBW-enhanced HRDA model and the reported results of HRDA \cite{hoyer2022hrda}. The vertical axis corresponds to the ground truth labels, while the horizontal axis represents the predicted classes. Diagonal is depicted with bold colors for visualization. Classes are sorted from least to most number of samples.}
        \label{fig:confmatrix}
    \end{subfigure}\hfill
    \begin{subfigure}[b]{0.48\linewidth}
        \includegraphics[trim=0 0 0 0,clip, width=1.1\linewidth,]{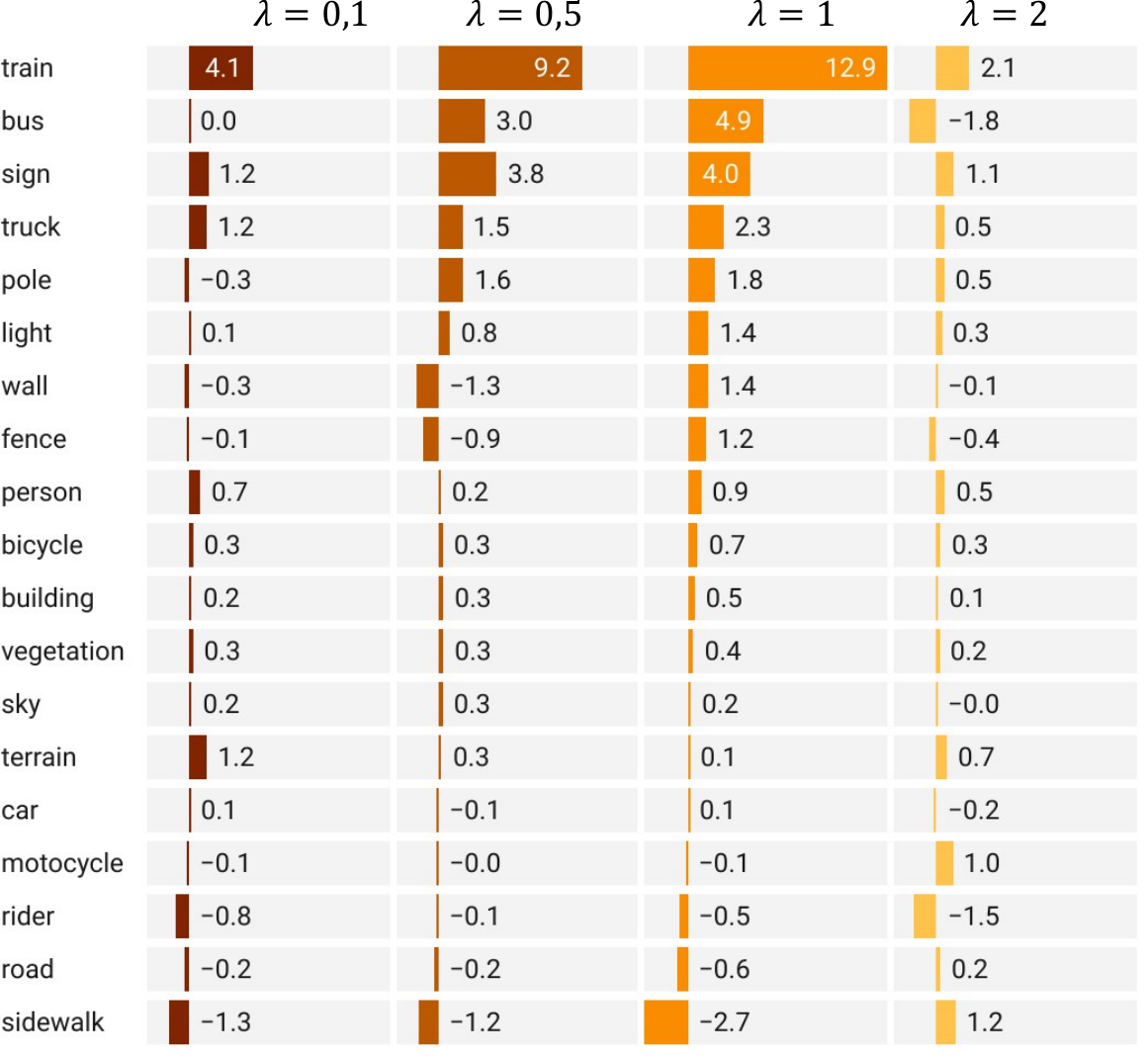}
        \caption{Per-class gain/downgrade compared to the baseline (i.e., without incorporating GBW) using different $\lambda$ values. Change is measured in mIoU points. Results for the GTA-Cityscapes scenario for HRDA method \cite{hoyer2022hrda}.}
        \label{fig:radar_ablation}
    \end{subfigure}
    \caption{GBW per-class performance analysis. }
    \vspace{-5mm}
\end{figure}

Figure \ref{fig:confmatrix}, depicts the disparity between the state-of-the-art model HRDA confusion matrix \cite{hoyer2022hrda} and the confusion matrix of the modified HRDA after the inclusion of GBW. Introducing GBW notably enhances recall, as evidenced by the predominantly positive values along the diagonal (15 out of 19). Note that the gain in recall is specially significant for the least represented classes in the source dataset (GTA): \textit{train, motocycle} and \textit{bicycle}. This is evidenced by the true positives on the diagonal. Although GBW-enhanced model shows  improvements for most classes, it is observed that some classes as \textit{sidewalk} and   \textit{motorcycle} do not benefit from it. This is further analyzed in the Supplementary material. %Even for classes where the mIoU decreases, the recall shows improvement.

% \begin{figure}[]
%     \centering
%     \includegraphics[width=.9\linewidth,]{HRDA_per_class.png}\\
%     %\includegraphics[width=.8\linewidth,]{HRDA_per_class_bars.png}
%     \caption{Comparison of per-class performance between our HRDA model (blue) and the reported results (red) \cite{hoyer2022hrda}. The radar plot illustrates the class mIoU values, with each axis representing a specific class.}
%     \label{fig:per-class}
%     %\vspace{-5mm}
% \end{figure}

%(Al material suplementario esta frase y la imagen) 

\begin{figure}[t] 
    % \begin{subfigure}[b]{0.45\linewidth}
    %     \includegraphics[width=\linewidth,]{frankfurt_000000_005543_gtFine_color.png}
    % \end{subfigure}
    % \begin{subfigure}[b]{0.45\linewidth}
    %     \includegraphics[width=\linewidth,]{frankfurt_000000_005543_leftImg8bit_Color.png}
    % \end{subfigure} \\
    % \begin{subfigure}[b]{0.3\linewidth}
    %     \includegraphics[trim=1460 550 270 250,clip, width=\linewidth,]{frankfurt_000000_005543_gtFine_color.png}
    % \end{subfigure}
    % \begin{subfigure}[b]{0.3\linewidth}
    %     \includegraphics[trim=1460 550 270 250,clip, width=\linewidth,]{frankfurt_000000_005543_leftImg8bit_LOW.png}
    % \end{subfigure}
    % \begin{subfigure}[b]{0.3\linewidth}
    %     \includegraphics[trim=1460 550 270 250,clip, width=\linewidth,]{frankfurt_000000_005543_leftImg8bit.png}
    % \end{subfigure}\\
    \begin{subfigure}[b]{0.246\linewidth}
        \includegraphics[trim=1060 270 60 100,clip, width=\linewidth,]{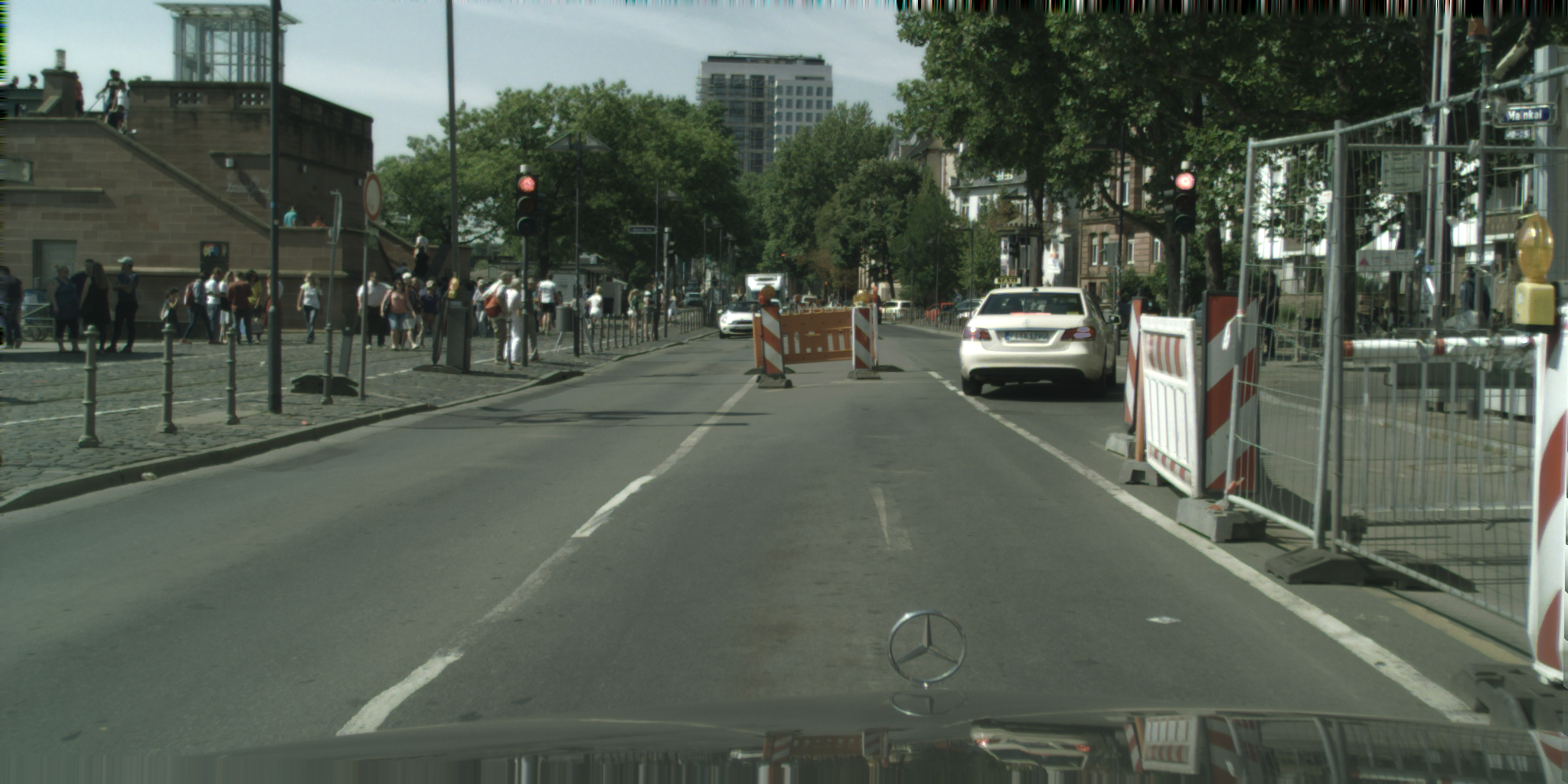}
    \end{subfigure}\hfill
    % \begin{subfigure}[b]{0.2\linewidth}
    %     \includegraphics[width=\linewidth,]{images/frankfurt_000001_017101_leftImg8bit_Color.png}
    % \end{subfigure} \hfill
    \begin{subfigure}[b]{0.246\linewidth}
        \includegraphics[trim=1060 270 60 100,clip, width=\linewidth,]{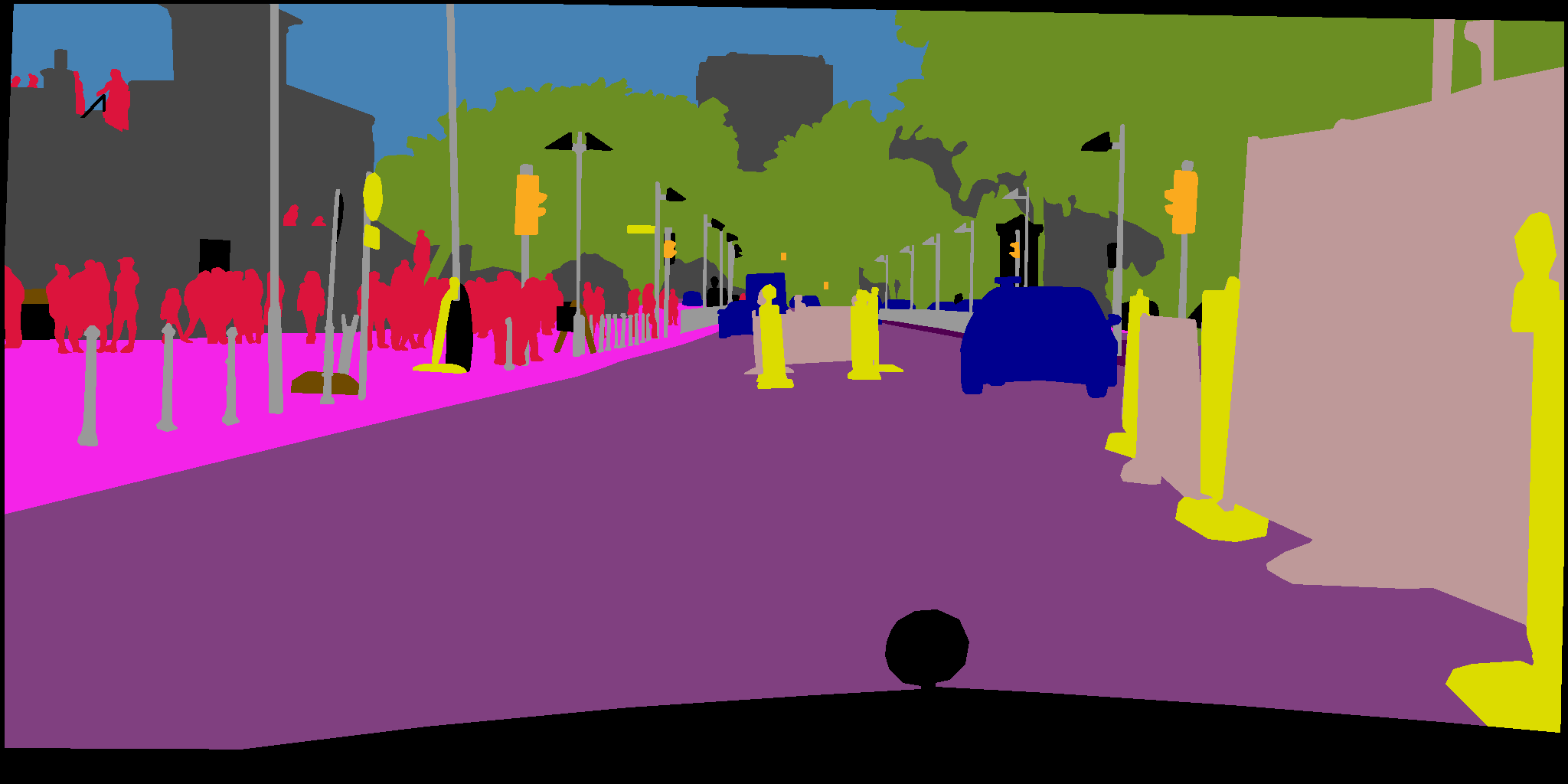}
    \end{subfigure}\hfill
    \begin{subfigure}[b]{0.246\linewidth}
        \includegraphics[trim=1060 270 60 100,clip, width=\linewidth,]{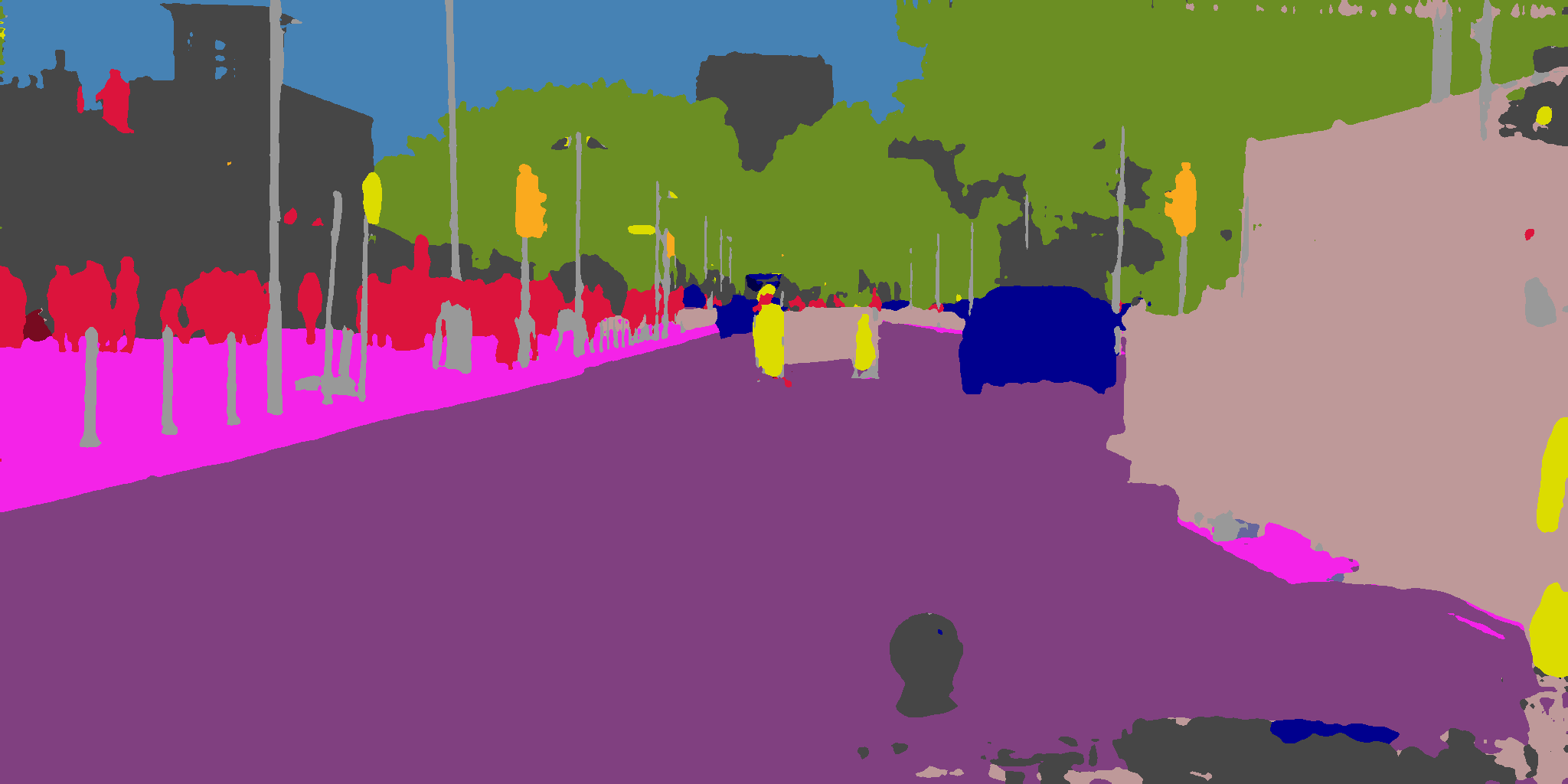}
    \end{subfigure}\hfill
    \begin{subfigure}[b]{0.246\linewidth}
        \includegraphics[trim=1060 270 60 100,clip, width=\linewidth,]{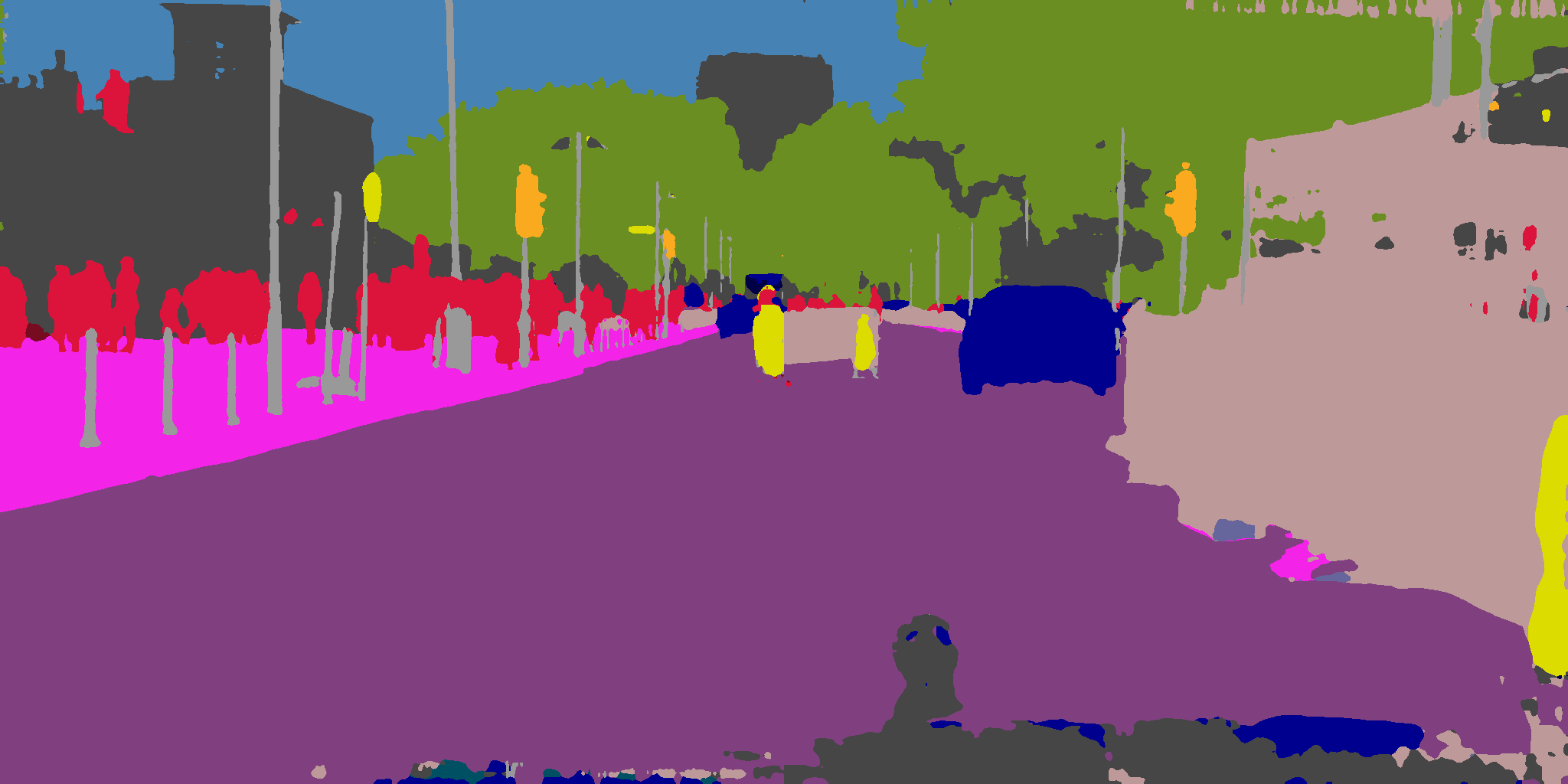}
    \end{subfigure}\\
    
    % \begin{subfigure}[b]{0.5\linewidth}
    %     \includegraphics[trim=10 100 870 1300,clip,width=\linewidth,]{images/munster_000134_000019_leftImg8bit.png}
    % \end{subfigure}\hfill
    \begin{subfigure}[b]{0.246\linewidth}
        \includegraphics[trim=10 1350 1260 150,clip,width=\linewidth,]{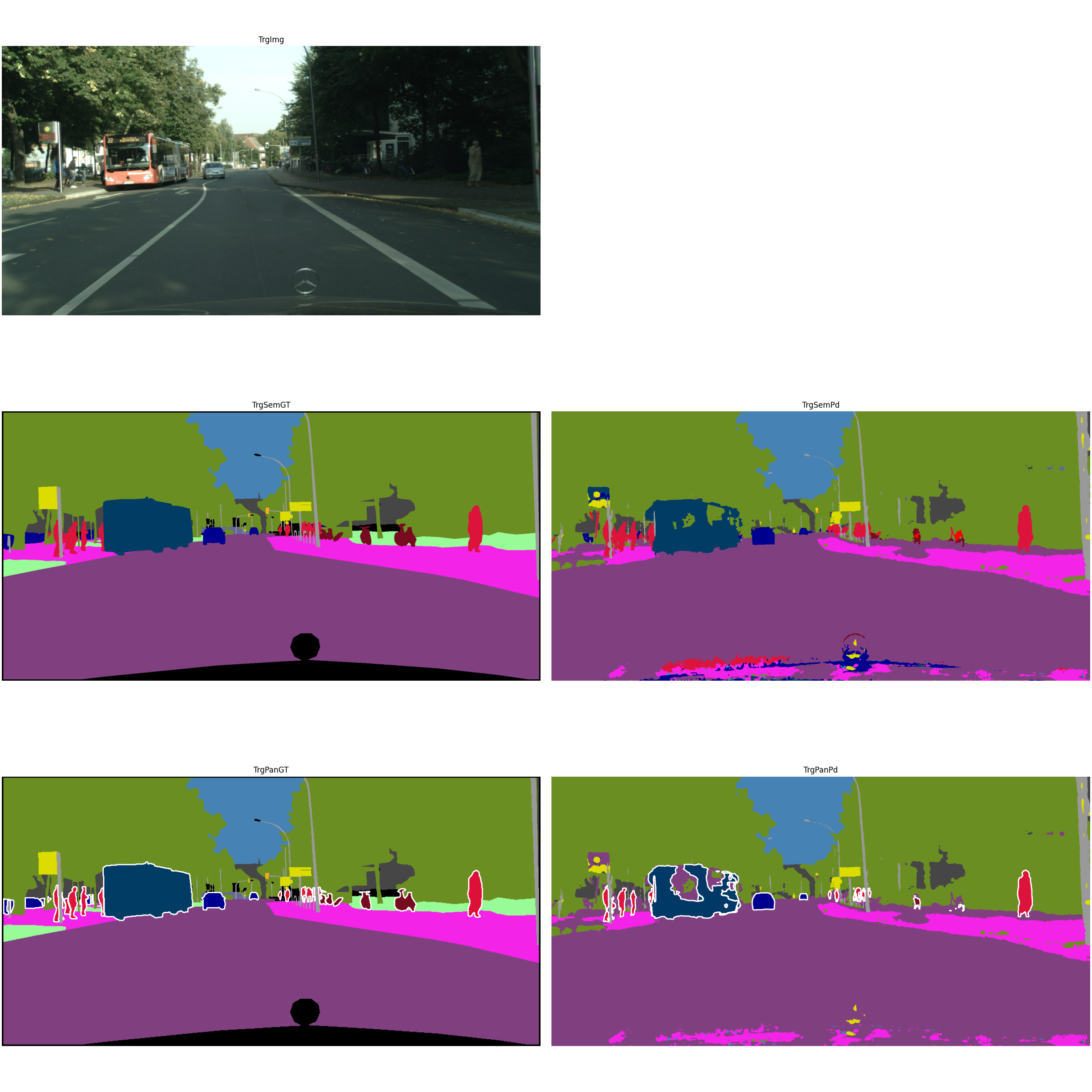}
    \end{subfigure}\hfill
    \begin{subfigure}[b]{0.246\linewidth}
        \includegraphics[trim=10 200 1260 1300,clip,width=\linewidth,]{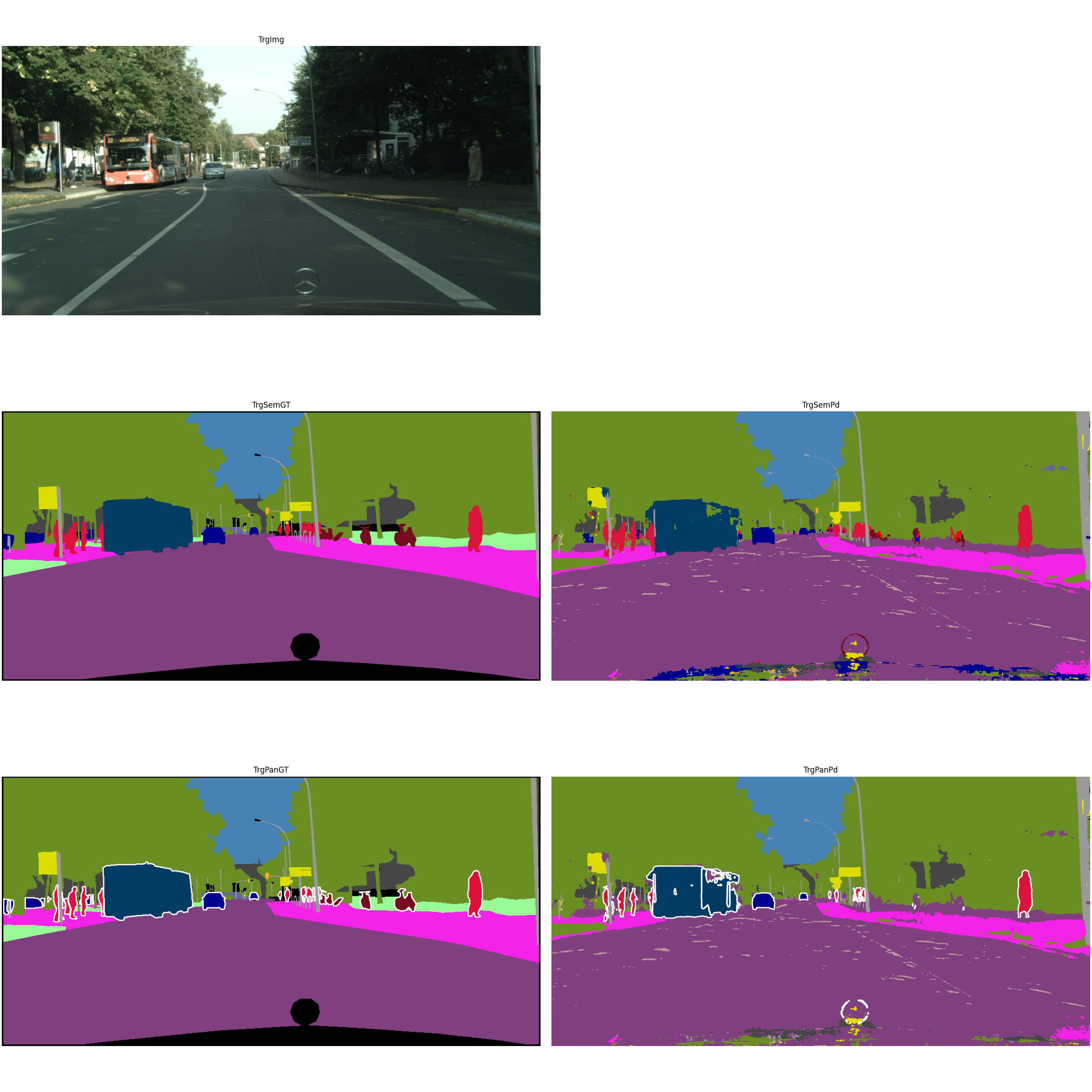}
    \end{subfigure}\hfill
    \begin{subfigure}[b]{0.246\linewidth}
        \includegraphics[trim=870 200 400 1300,clip,width=\linewidth,]{images/munster_000134_000019_leftImg8bit_LOW.png}
    \end{subfigure}\hfill
    \begin{subfigure}[b]{0.246\linewidth}
        \includegraphics[trim=870 200 400 1300,clip,width=\linewidth,]{images/munster_000134_000019_leftImg8bit.png}
    \end{subfigure}
    \caption{Qualitative comparison. Column-wise: image, ground truth, model trained with and without GBW. Semantic segmentation ($1^{st}$ row). Panoptic segmentation ($2^{nd}$ row). }
    \vspace{-5mm}
    \label{fig:comparison}
\end{figure}

Figure \ref{fig:comparison} includes a qualitative comparison between models trained with and without GBW. The GBW-model exhibits enhanced capability in accurately classifying less populated classes. See examples for \textit{fence} (second row), \textit{bus} and \textit{sign} (fourth row). See Supplementary material for more visual examples. %However, it also reveals that our model presents a worse performance for instances of different annotation pattern, as exemplified by the sidewalk class in the second example.

\paragraph{Regularization weight}
This section examines the impact of the regularization factor when GBW is incorporated into the HRDA model \cite{hoyer2022hrda} for semantic segmentation, as shown in Table \ref{tab:ablation}. Note that $\lambda$ represents the weight of the regularization term in Equation \ref{eq:lambda}, with higher values enforcing uniform weights, which is equivalent to not applying any weighting at all. This is clearly observed in the performance when $\lambda=10$, which matches the performance of a model trained without GBW. Conversely, extremely small $\lambda$ values may cause the weights of certain classes approach to zero, effectively disregarding them during training. This accounts for the suboptimal performance observed when $\lambda=0.01$. Figure \ref{fig:radar_ablation} visualize disentangles this effect in per-class performance. GBW drastically affects \textit{train} and visually similar classes (\textit{bus}, \textit{truck}) performance, with improvements of over +10 IoU. We believe this to be because \textit{train} is the second least represented class in the GTA \cite{Richter_2016_ECCV} dataset and models trained on GTA tend to classify it as \textit{bus} and \textit{truck}, thus improving also their IoU by reducing FPs. %These results showcase the impact of GBW by improving the models recall on under-represented categories and false positives of visually similar classes of those classes.

\begin{table}[] 
    \centering
    \vspace{-3mm}
    \setlength{\tabcolsep}{15pt}
     \resizebox{\linewidth}{!}{%
    \begin{tabular}{l c c c c c c c}
    $\lambda$ & - & 0.01 & 0.1 & 0.5 &1 & 2 & 10\\\toprule
    mIoU & 73.2& 56 & 73.5 & 73.8&74.7&73.8&73.2\\
    Gain & -& -16.8&0.3&0.6&1.5&0.6&0.0\\\bottomrule
    
    \end{tabular}}
%     \begin{tabular}{c c c c}
%          %Architecture
%           $\lambda$ & mIoU & Gain\\\toprule
%           - & 73.2 & \\
%           0.01 & 56& -16.8\\
%           0.1 & 73.5 & 0.3\\
%           0.5 & 73.8 & 0.6\\
%           1 & 74.7& 1.5\\
%           2 & 73.8& 0.6\\
%           10 & 73.2& 0.0\\\bottomrule
% %         Ours& 0.01& &\\
% %         Ours& 0.001& &\\\midrule
%     \end{tabular}
    
    \caption{Ablation study for the $\lambda$ weight on the GTA-Cityscapes framework embedded into the state-of-the-art UDA method HRDA\cite{hoyer2022hrda}}
    \label{tab:ablation}
    \vspace{-13mm}
\end{table}

% \begin{figure}[]
%     \centering
%     \includegraphics[width=.75\linewidth]{images/AblationLow.pdf}
%     \caption{Per-class IoU difference of the models trained with different $\lambda$ weights and uniform weights. Red area represents no improvement or drop in performance compared to using no weights. The further from the red area represents a bigger relative improvement. Performance is measured in mIoU, results on the GTA-Cityscapes framework for HRDA method \cite{hoyer2022hrda}.}
%     \label{fig:radar_ablation}
%     \vspace{-2mm}
% \end{figure}

\begin{figure}[t]
    \centering
    \begin{subfigure}[b]{.76\linewidth}
    \includegraphics[width=\linewidth]{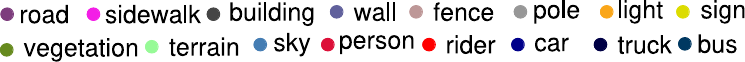}
    \end{subfigure}\hfill
    \begin{subfigure}[b]{0.45\linewidth}
    \centering
    \includegraphics[width=.75\linewidth]{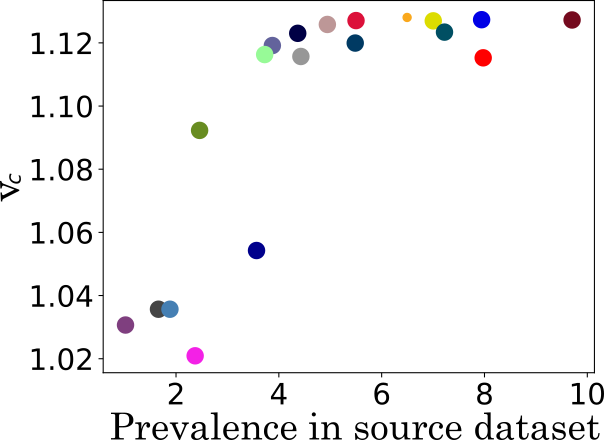}
    \caption{Visual comparison between class prevalence and GBW maximum assigned weights. Prevalence measured by negative log percentage of pixels in the source dataset.}
    \label{fig:prevvsweight}
    \end{subfigure}\hfill
    \begin{subfigure}[b]{0.45\linewidth}
    \centering
    \includegraphics[width=.75\linewidth]{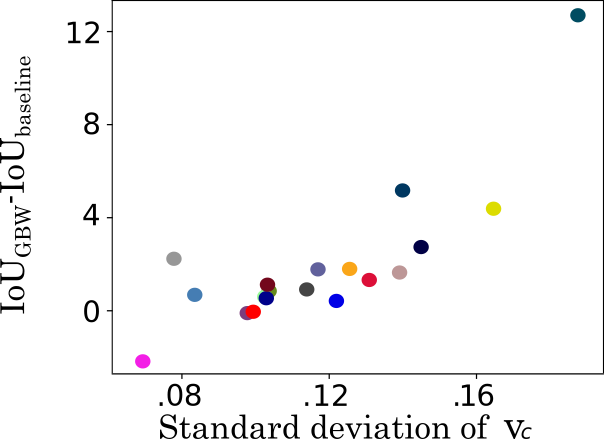}
    \caption{Relationship between performance improvements of GBW and the standard deviation of the GBW assigned during training.\\}
    \label{fig:stdvsiou}
    \end{subfigure}\hfill
    \caption{Analysis of the GBW weights assigned throughout the training with HRDA \cite{hoyer2022hrda}. }
    \vspace{-3mm}
\end{figure}

\paragraph{Weight distribution}
The main hypothesis is that a GBW-enhanced model tends to promote the learning of under-represented classes. To visually inspect this, Figure \ref{fig:prevvsweight} compares the percentage of pixels per class in the source dataset with the maximum weight assigned by GBW, showcasing that highly represented classes tend to be given a smaller weight than less represented classes. The Figure only depicts the maximum weight per class along the training, as weights are made to be very variable between training iterations depending on the current state of the model. It can be observed that classes that presented a large variation in weights, such as \textit{train} and  \textit{sign}, are associated with larger improvements in performance, with increases of over 12 points in mIoU. Figure \ref{fig:stdvsiou} illustrates the relationship between weight variance and mIoU. Note how classes with performance improvements highly correlate with the variability of the assigned weights.  Additionally, highly represented classes as \textit{road} and \textit{sidewalk}, exhibit more uniform weights, indicated by a low standard deviation.
\paragraph{Weight evolution}
Figures \ref{subfig_Coarse} and \ref{subfig_ped} present the averaged evolution of the per-class weights assigned by GBW. Coarse and high populated classes (as shown in Figure \ref{subfig_Coarse}) start the training process with higher weights. However, as training progresses, these weights gradually decrease and are distributed to other classes. Specifically, looking at two similar classes: \textit{person} (highly populated) and \textit{rider} (under-populated) GBW adeptly handles them by assigning higher weights to \textit{person} at the onset of training, while lowering the weight for \textit{rider} (see Figure \ref{subfig_ped}). As the training proceeds, the weight assigned to \textit{rider} gradually increases, while decreasing simultaneously the weight for \textit{person} suggesting an interesting capacity of GBW in adapting to the training outcomes. This, together with previous evidences, suggests that the success of GBW relies on the algorithm's capacity to dynamically assign appropriate weights to each class at different stages of the learning process naturally and dynamically integrating evidences from all classes.

\section{Conclusion}
Despite the extensive literature in UDA, the impact of statistical discrepancies\textemdash such as different class imbalances between the source and target dataset\textemdash is still an open topic. Due to the lack of target labels, algorithmic-level techniques to handle class imbalance are deemed unsuited for UDA in dense prediction visual tasks. In this work, class imbalance is handled by an adaptive weighting scheme based on the loss gradients at each training step. Per-class weights are updated on-the-fly at each training step to compensate for the magnitudes of the loss gradient. Experimental results show improvements in the recall of scarcely represented classes. The benefits of GBW are exhaustively explored by means of its introduction into six different state-of-the-art UDA methods for semantic segmentation and panoptic segmentation. The results consistently show that incorporating GBW leads to performance gains across different architectures and datasets for both semantic and panoptic segmentation. We hope that GBW can be used on future UDA methods by serving as a building block for narrowing the gap between UDA and supervised methods.

%Ablation studies examine the impact of our per-class weighting and sampling techniques on each individual class performance. The results showed that incorporating both per-class weights and sampling techniques led to significant improvements in model performance.  

%\clearpage  % TODO REVIEW/FINAL: This \clearpage needs to be removed from both review and camera-ready versions.

\bibliographystyle{splncs04}
\bibliography{main}
\end{document}